%% file: main.tex
\documentclass[10pt,twocolumn,letterpaper]{article}

\usepackage{wacv}              % To produce the CAMERA-READY version

\usepackage[accsupp]{axessibility}

\usepackage[utf8]{inputenc} % allow utf-8 input
\usepackage[T1]{fontenc}    % use 8-bit T1 fonts
\usepackage{booktabs}
\usepackage{amsfonts}       % blackboard math symbols
\usepackage{amsmath}
\usepackage{amssymb}
\usepackage{nicefrac}       % compact symbols for 1/2, etc.
\usepackage{microtype}      % microtypography
\usepackage{xcolor}         % colors

\usepackage{times}
\usepackage{epsfig}
\usepackage{graphicx}

\usepackage{multicol}
\usepackage{multirow}
\usepackage{tabularx}
\usepackage{subcaption}

\makeatletter
\robustify\@latex@warning@no@line
\makeatother
\usepackage[noblocks]{authblk}

\usepackage[title]{appendix}

\usepackage[pagebackref,breaklinks,colorlinks]{hyperref}
\usepackage{verbatim}

\newcommand{\appendixhead}%
{\centering\textbf{\huge Appendix}
\vspace{0.25in}}

\makeatletter
\renewcommand\AB@affilsepx{, \protect\Affilfont}
\makeatother
\usepackage[capitalize]{cleveref}
\crefname{section}{Sec.}{Secs.}
\Crefname{section}{Section}{Sections}
\Crefname{table}{Table}{Tables}
\crefname{table}{Tab.}{Tabs.}

\def\wacvPaperID{0576} % *** Enter the WACV Paper ID here
\def\confName{WACV}
\def\confYear{2024}

\newcommand\gkd{GKD}
\newcommand\mkd{MKD}
\newcommand\tkd{TKD}
\newcommand\ta{TA}

\begin{document}

\title{\vspace{-1.8cm}Adapt Your Teacher: Improving Knowledge Distillation for Exemplar-free Continual Learning}

\author[1,2]{\vspace{-1cm}Filip~Szatkowski\thanks{corresponding author, email: \href{mailto:filip.szatkowski.dokt@pw.edu.pl}{filip.szatkowski.dokt@pw.edu.pl}}}
\author[2,3,4]{Mateusz~Pyla}
\author[2,3,4]{Marcin~Przewięźlikowski}
\author[2,5]{\\Sebastian~Cygert}
\author[2,6,7]{Bartłomiej~Twardowski}
\author[1,2,8]{Tomasz~Trzciński}
\affil[1]{\normalsize Warsaw~University~of~Technology}
\affil[2]{IDEAS~NCBR}
\affil[3]{Jagiellonian~University,~Faculty~of~Mathematics~and~Computer~Science}
\affil[4]{Jagiellonian~University,~Doctoral~School~of~Exact~and~Natural~Sciences}
\affil[5]{Gdańsk~University~of~Technology}
\affil[6]{Autonomous~University of~Barcelona}
\affil[7]{Computer~Vision~Center}
\affil[8]{Tooploox\vspace{-0.5cm}}

\maketitle

%%%%%%%%% ABSTRACT

\begin{abstract}
\vspace{-0.3cm}
In this work, we investigate exemplar-free class incremental learning (CIL) with knowledge distillation (KD) as a regularization strategy, aiming to prevent forgetting. KD-based methods are successfully used in CIL, but they often struggle to regularize the model without access to exemplars of the training data from previous tasks. Our analysis reveals that this issue originates from substantial representation shifts in the teacher network when dealing with out-of-distribution data. This causes large errors in the KD loss component, leading to performance degradation in CIL models. Inspired by recent test-time adaptation methods, we introduce Teacher Adaptation (TA), a method that concurrently updates the teacher and the main models during incremental training. Our method seamlessly integrates with KD-based CIL approaches and allows for consistent enhancement of their performance across multiple exemplar-free CIL benchmarks. The source code for our method is available at \href{https://github.com/fszatkowski/cl-teacher-adaptation}{https://github.com/fszatkowski/cl-teacher-adaptation}.
\vspace{-0.3cm}

\end{abstract}

%%%%%%%%% BODY TEXT
\section{\vspace{-0.1cm}Introduction}
\label{sec:intro}

% About CL
Continual learning aims to create machine learning models capable of acquiring new knowledge and adapting to evolving data distributions over time.
One of the most challenging continual learning scenarios is \emph{class incremental learning} (CIL)~\cite{van2019three,masana2022class}, where the model is trained to classify objects incrementally from the sequence of tasks, without forgetting the previously learned ones. 

\begin{figure}[t]
    \centering
    \includegraphics[width=0.9\columnwidth]{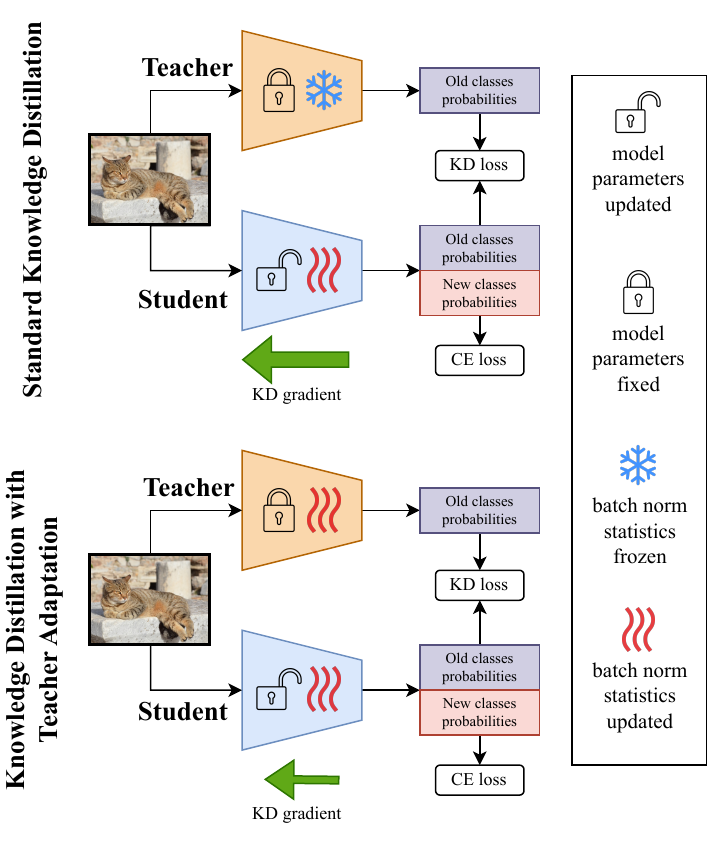}

   \caption{Enhancement of vanilla Knowledge Distillation approach used in Continual Learning with our method of Teacher Adaptation. When training the student model on the new task, we allow the teacher model to continuously update its batch normalization statistics, which reduces the divergence between the representations in both models. Our method leads to lower knowledge distillation loss and an overall more stable model.\vspace{-0.5cm}}
   \label{fig:method}
\end{figure}

A simple and effective method of reducing forgetting is by leveraging \emph{exemplars}~\cite{rebuffi2017icarl,iscen2022memory,bang2021rainbow,prabhu2020gdumb} of previously encountered training examples, {\it e.g.} by replaying them or using them for regularization.
However, this approach presents challenges, particularly in terms of high storage needs and privacy concerns. These problems can affect edge devices, due to their limited storage capacity, and medical data, given their sensitive nature. 

% 1 x 3 teaser
\begin{figure*}[t]
    \centering
      \begin{subfigure}[b]{0.33\textwidth}
        \centering
        \includegraphics[trim=10 0 10 0 ,clip,width=\textwidth]{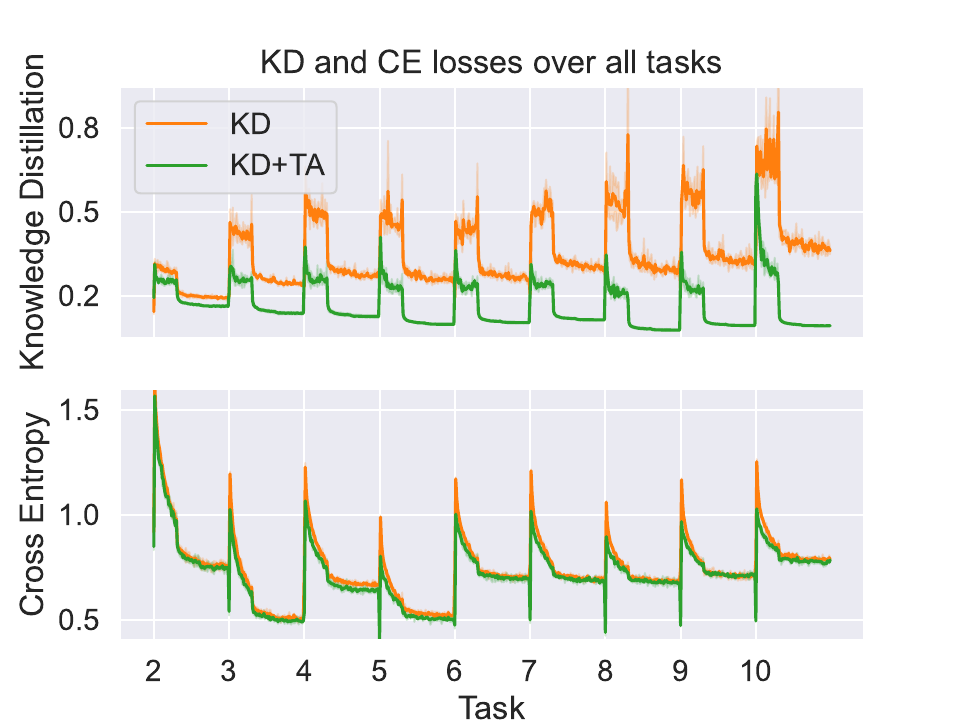}
          \end{subfigure}
    \begin{subfigure}[b]{0.33\textwidth}
        \centering
        \includegraphics[trim=10 0 10 0 ,clip,width=\textwidth]{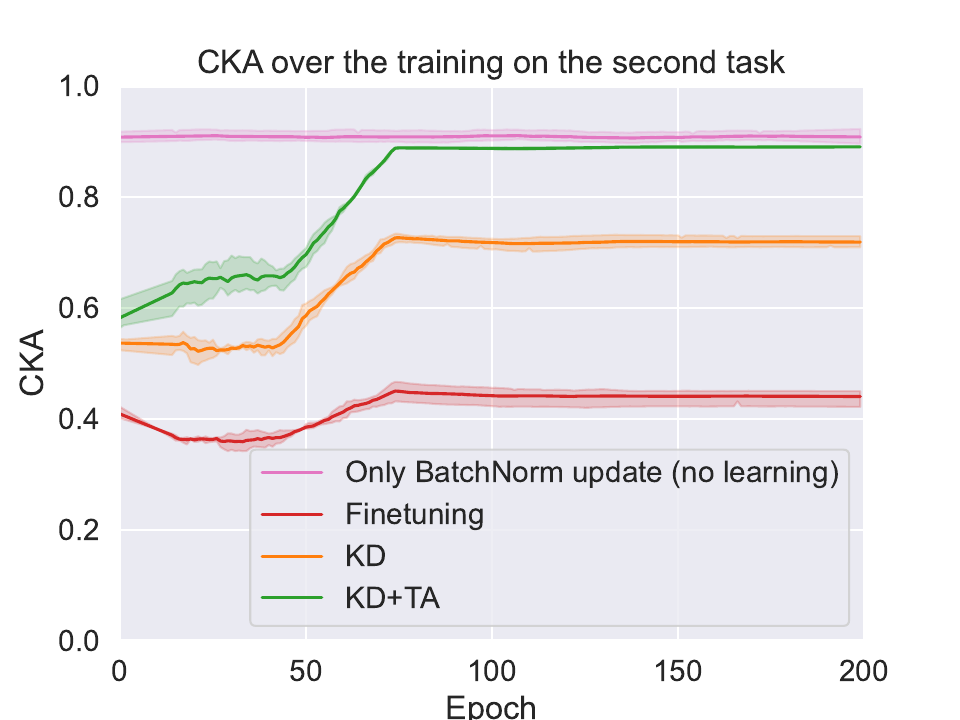}
    \end{subfigure}
    \begin{subfigure}[b]{0.33\textwidth}
        \centering
        \includegraphics[trim=10 0 10 0 ,clip,width=\textwidth]{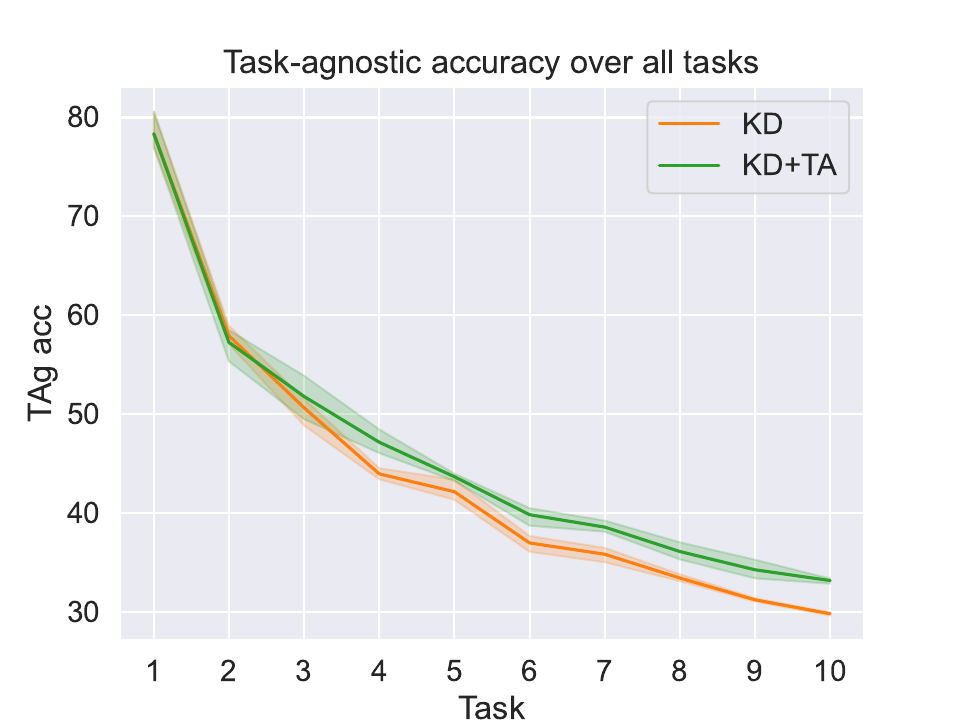}
    \end{subfigure}

   \caption{
   Applying our teacher adaptation (TA) method reduces knowledge distillation (KD) loss and improves stability throughout continual learning. 
 (left) KD loss and cross-entropy (CE) loss of training the model with and without TA. Our method leads to more consistent representation, as visualized by the CKA~\cite{kornblith2019similarity} between the representations of the new data obtained in the teacher and student models while learning the second task (middle). KD with \ta\ leads to better task-agnostic accuracy (right). We conduct the experiments on CIFAR100 split into 10 tasks.
}
   \label{fig:teaser}
\end{figure*}

A common approach for exemplar-free CIL is knowledge distillation (KD), where the current model (student) is trained on the new data with a regularization term that minimizes the output difference with the previous model (teacher), which is kept frozen. This approach was introduced by LwF~\cite{li2017learning} and has been extended by many other methods. However, most of these methods use exemplars, such as iCaRL~\cite{rebuffi2017icarl}, EEIL~\cite{castro2018end}, LUCIR~\cite{hou2019learning}, PodNET~\cite{douillard2020podnet}, SSIL~\cite{ahn2021ss}, or rely on external datasets~\cite{dmc,lee2019overcoming}.

Exemplar-free CIL still remains challenging~\cite{smith2023closer} for KD methods due to the possibility of significant distribution drift in subsequent tasks. Such drift leads to large errors during training with KD loss, causing more undesired changes in the main model and harming the overall performance of the CIL training. This raises the question: \textit{Can we adjust the teacher model to better transfer knowledge from earlier tasks?}

Motivated by the recent domain adaptation methods~\cite{TENT,steffen2020shiftadapt}, we examine the role of batch normalization statistics in CIL training. We conjecture that in standard KD methods, the KD loss between models with different normalization statistics may introduce unwanted model updates due to the data distribution shifts. To avoid this, we propose to continuously adapt them to the new data for the teacher model while training the student.

We show that adapting the teacher's batch normalization statistics to the new task can significantly lower KD loss without affecting the CE loss, which reduces changes in the model's representations (\Cref{fig:teaser}). 
We note that, while the idea of changing the teacher model was explored in the standard KD settings~\cite{zhou2022BERTteach,ma2022knowledge}, our approach is the first application of this idea to CIL scenario, where the teacher and the model are trained on non-overlapping data. Moreover, our method works differently by exploiting the batch normalization statistics.
We apply our method on top of different distillation strategies in CIL and show consistent improvements across various settings.

In summary, we make the following contributions: 
\begin{enumerate}
    \item We revisit the KD-based class-incremental learning (CIL) framework and study the negative impact of regularization using out-of-distribution data. We are the first to highlight the need for adjusting the teacher model in an exemplar-free situation, where it is usually kept frozen.
    \item We propose a simple yet highly effective technique called Teacher Adaptation (TA), that enhances KD for exemplar-free CIL.
    \item Through extensive experiments, we demonstrate that \ta\ can be seamlessly integrated with various KD approaches, leading to significant improvements over the baselines across a wide range of continual learning scenarios for various datasets. We show that those improvements hold even when using pretrained models or in the presence of substantial distributional shifts between consecutive domains.
\end{enumerate}

\section{Related works}

\textbf{Class Incremental Learning (CIL)}~\cite{van2019three,masana2022class} is a subfield of continual learning, where the aim is to learn incrementally from a stream of tasks, without the task identifier. 
There exist several families of approaches to CIL:

Memory-based methods store either exemplars or features from the previous tasks in the buffer~\cite{rebuffi2017icarl,iscen2022memory,bang2021rainbow,prabhu2020gdumb} and use them during training the new task to consolidate previously learned knowledge. Those methods usually perform well, but their practical applications are limited due to privacy concerns and memory requirements that arise when storing the data. Architectural approaches focus on modifying the structure of the model, often by allocating certain parameters to corresponding tasks~\cite{wang2022learning,wang2022dualprompt}. Finally, regularization-based methods aim to preserve the knowledge in the network by imposing constraints on the changes in model weights~\cite{kirkpatrick2017overcoming} or activations during learning the new task~\cite{li2017learning}. Many CIL methods often also combine the above approaches~\cite{castro2018end,Wu_2019_CVPR,rolnick2019experience}, for instance using both memory and regularizaion~\cite{rebuffi2017icarl,li2017learning,ahn2021ss}.

\textbf{Regularization methods for continual learning} offer a way to prevent forgetting with constant memory usage and no privacy issues. There are two main types of regularization methods: (i) parameter regularization and (ii) functional regularization. The first type of methods regularizes the model weights, for example using the Fisher Information Matrix~\cite{kirkpatrick2017overcoming}, synaptic saliency~\cite{zenke2017continual} or the gradient inspection~\cite{aljundi2018memory}. On the other hand, functional regularization methods employ knowledge distillation (KD) techniques to regularize model activations. KD was originally proposed by Hinton et al.~\cite{hinton2015distilling} to transfer the knowledge from a larger model to a smaller one. In CL, KD was first applied in Learning without Forgetting (LwF)~\cite{li2017learning}, where the model is discouraged from drifting too far from the model from previous tasks. We describe KD methods in detail in~\Cref{sec:kd_in_cl}.

\textbf{Functional regularization} has been widely used in CL since the introduction of LwF~\cite{rebuffi2018efficient,hou2019learning,rebuffi2017icarl,prabhu2020gdumb} and numerous variants of KD have been proposed. Particularly, SSIL~\cite{ahn2021ss} uses task-wise knowledge distillation, while PODNET~\cite{douillard2020podnet} regularizes using spatial-based distillation loss applied throughout the model. 

Multi-level knowledge distillation~\cite{ding2021multilevel} uses the current model to distill the knowledge from pruned snapshots of all previous models, while ANCL~\cite{kim2023achieving} distills simultaneously from the previous task model and the model learned specifically for the new task. Moreover, DMC~\cite{zhang2020classincremental} uses knowledge distillation on an auxiliary dataset to consolidate the knowledge from previous tasks.

However, most of those methods use a memory buffer and their performance depends on it heavily, which makes them impractical for exemplar-free settings. 
Recently, several works explored the idea of modifying the teacher model through meta-learning for better knowledge transfer in standard KD setting~\cite{zhou2022BERTteach,ma2022knowledge}. Similarly, works on KD~\cite{heo2019comprehensive} suggest that updating the normalization statistics of the teacher model on the data used to train the student improves the performance. However, to our knowledge, our method is the first one that explores different approaches to teacher adaptation, such as updating the normalization statistics, in the context of CIL.

\textbf{Batch Normalization} (BN)~\cite{ioffe2015batch} is widely used in deep learning models, but can be problematic for settings where the data distribution changes over time. Alternative approaches such as LayerNorm~\cite{ba2016layer} or GroupNorm~\cite{wu2018group} do not rely on the batch-wise statistics, but directly replacing BN layers with them was shown to often decrease the performance of the models.
Several domain adaptation methods achieve domain transfer through the use of normalization statistics~\cite{TENT,steffen2020shiftadapt}. Recent work on efficient finetuning of large language models using only normalization layers ~\cite{qi2022parameter} also suggests that the normalization layers play a crucial role in training deep neural networks.
In CL, it was shown that BN can cause a discrepancy between the training and testing phases of BN, as the testing data is normalized using the statistics biased towards the current task, which results in higher forgetting of older tasks~\cite{santurkar2019does}. Several works have attempted to address this issue by CL-specific modifications to BN~\cite{pham2022continual,cha2023rebalancing}. However, those approaches are not suited for exemplar-free settings. 

\section{Method}
\label{sec:method}

In class-incremental learning setup, the model learns tasks sequentially. Each task contains several classes which are disjoint with the classes in other tasks. During training task $t$, we only have access to the data $D^t$ from task $t$ which contains images $\mathbf{x}_{i} \in X^t$ with class labels $y_i \in C^t$. Thus an incremental learning problem $\mathcal{T}$ with $n$ tasks can be formulated as: 
$\mathcal{T}=\left \{ D^1, D^2,...,D^t,..., D^n \right \}$,
%=\left \{ \left ( X^1, C^1 \right ) ,\left ( X^2,C^2 \right ),...,\left ( X^t,C^t \right ),...(X^n, C^n)\right \}$. 
where after training $n$ tasks we evaluate the model on all classes $ C^{1} \cup \ldots \cup C^n$, without knowing the task label at inference time (this is different than task-incremental learning, where task id can be used).

Below, we first introduce standard KD-based methods for exemplar-free CIL. Then we outline a problem of diverging batch normalization statistics between the teacher and student model caused by the shifts in training data between subsequent tasks. Finally, we propose to address this issue with a method that we call \emph{Teacher Adaptation} - a simple, yet effective solution that allows the teacher model to continuously update its normalization statistics alongside the student when training on the new data. The method in comparison to standard LwF is presented in \Cref{fig:method}.

\subsection{Knowledge Distillation in Continual Learning}
\label{sec:kd_in_cl}

Knowledge distillation is one of the most popular techniques employed to reduce forgetting between subsequent tasks in incremental learning. Continual learning methods that use knowledge distillation save the model $\Theta_{t}$ (\emph{teacher}) trained after each task $t$ and use it during learning the model $\Theta_{t+1}$ (\emph{student}) on new task $t+1$. The learning objective for task $t+1$ then becomes:
\begin{equation}
    L = L_{CE} + \lambda L_{KD},
\end{equation}
where $L_{CE}$ is the cross-entropy loss for classification on new data, $L_{KD}$ is the knowledge distillation loss computed using $\Theta_{t}$ and $\Theta_{t+1}$, and $\lambda$ is the coefficient that controls the trade-off between stability and plasticity. The general formula for knowledge distillation loss can include either output from the final layer of the model~\cite{li2017learning,rebuffi2017icarl,ahn2021ss}, or also representations from intermediate model layers~\cite{dhar2019learning,douillard2020podnet}. In practice, most exemplar-free methods that use knowledge distillation compute knowledge distillation loss using only the final layer outputs, and various methods that use intermediate representations usually only perform well with exemplars~\cite{smith2023closer}.

Multiple variants of knowledge distillation loss were proposed for continual learning. In exemplar-free CIL, KD loss is usually computed with the logits $y_{o}$, $\hat{y_{o}}$ returned by the current and previous models respectively. Following Li et al.~\cite{li2017learning}, we denote that the loss uses logits corresponding to previously seen classes with a subscript $o$. Ahn et al.~\cite{ahn2021ss} classify KD methods into general KD (GKD), which aggregates together logits belonging to the classes from all the previous tasks, and task-wise KD (TKD), which treats classes within each task separately. 

GKD loss appears in several works~\cite{lee2019overcoming,wu2019large,zhao2020maintaining} and usually uses cross-entropy:
\begin{align}
\label{eq:gkd_loss}
    \mathcal{L}_{GKD}(\mathbf{y}_o,\mathbf{\hat{y}}_o) &= - \sum_{i=1}^{|C_{t}|} p_o^{(i)} \log \hat{p}_o^{(i)},
\end{align}
where $p_o^{(i)}$ is the probability of the $i$-th class and $|C_{t}|$ is the number of classes learned by previous model $\Theta_{t}$. Probabilities $p_o^{(i)}$, $\hat{p}_o^{(i)}$ are computed with temperature parameter $T$ as follows:
\begin{equation}
\label{eq:temp_norm}
p_o^{(i)} = \dfrac{e^{y_o / T}}{\sum_j e^{y_o / T}}, \quad\hat{p_o}^{(i)} = \dfrac{e^{\hat{y_o} / T}}{\sum_j e^{\hat{y_o} / T}}
\end{equation}

Comparatively, TKD loss, which was also used in several works~\cite{li2017learning,castro2018end,ahn2021ss}, sums of the separately computed losses for each task:
\begin{align}
\label{eq:tkd_loss}
    \mathcal{L}_{TKD}(\mathbf{y}_o,\mathbf{\hat{y}}_o) &= \sum_{i=1}^{t} \mathcal{D}_{KL}(p_o^{(i)} \log \hat{p}_o^{(i)}),
\end{align}
where $\mathcal{D}_{KL}$ is Kullback–Leibler divergence and $p_o^{(i)}$, $\hat{p}_o^{(i)}$ are computed task-wise across the classes that belong to task $i$ as in \Cref{eq:temp_norm}.

Rebuffi et al.~\cite{rebuffi2017icarl} proposed another distinct variant of KD for multi-class incremental learning, where the loss is computed element-wise for each class (MKD):
\begin{align}
\label{eq:icarl_loss}
    \mathcal{L}_{\mkd\ }(\mathbf{y}_o,\mathbf{\hat{y}}_o) &= - \sum_{i=1}^{|C_{t}|} \sigma(y_o^{i}) \log \sigma(\hat{y}_o^{i}),
\end{align}
where $\sigma$ is a sigmoid function.

Additionally, a more recent KD-based method, Auxiliary Network Continual Learning (ANCL)~\cite{kim2023achieving}, explores the idea of multi-teacher KD for continual learning. ANCL trains an auxiliary network trained only for the current task and combines standard GKD loss with the KD loss computed between outputs for the current task and outputs of this auxiliary network.

In this work, we investigate the aforementioned KD techniques (\gkd, \mkd, \tkd, ANCL).

\subsection{Teacher Adaptation} 

Most models used in class incremental learning for vision tasks are convolutional neural networks such as ResNet~\cite{he2016deep}. Those models typically use batch normalization layers and keep the parameters and statistics of those layers in the teacher model $\Theta_{t}$ fixed during learning $\Theta_{t+1}$. However, with the changing distribution of the data in the new task, batch normalization statistics in the student and teacher models quickly diverge, which leads to higher KD loss. Gradient updates in this case not only regularize the model towards the stable performance on previous tasks but also compensate for the changes in normalization statistics, which needlessly overwrites the knowledge stored in the model and harms the learning process.

Inspired by test-time adaptation methods~\cite{TENT}, we propose to reduce this negative interference with a simple method that we label \emph{Teacher Adaptation (TA)}. Our method updates batch normalization statistics of both models simultaneously on new data while learning the new task. As shown in \Cref{fig:teaser}, it allows for significantly reduced KD loss over learning from sequential tasks in CIL, which improves the overall model stability. We provide additional analysis on why TA improves learning with KD in \Cref{sec:exp:bn_ablations} and in the Appendix.

\begin{table*}[!ht]
\centering
\resizebox{\textwidth}{!}{%
\begin{tabular}{@{}llrrrrrrrr@{}}
\toprule
 & &
  \multicolumn{4}{c}{Equal split of classes} &
  \multicolumn{4}{c}{Start from half of the classes} \\ \cmidrule{3-10}
 & &
  \multicolumn{2}{c}{10 tasks} &
  \multicolumn{2}{c}{20 tasks} &
  \multicolumn{2}{c}{11 tasks} &
  \multicolumn{2}{c}{26 tasks} \\ \cmidrule{3-10}
 & &
  \multicolumn{1}{c}{$Acc_{Inc} \uparrow$} &
  \multicolumn{1}{c}{$Forg_{Inc} \downarrow$} &
  \multicolumn{1}{c}{$Acc_{Inc} \uparrow$} &
  \multicolumn{1}{c}{$Forg_{Inc} \downarrow$} &
  \multicolumn{1}{c}{$Acc_{Inc} \uparrow$} &
  \multicolumn{1}{c}{$Forg_{Inc} \downarrow$} &
  \multicolumn{1}{c}{$Acc_{Inc} \uparrow$} &
  \multicolumn{1}{c}{$Forg_{Inc} \downarrow$}  \\ \midrule
 \parbox[t]{2.5mm}{\multirow{8}{*}{\rotatebox[origin=c]{90}{a) CIFAR100}}} & \gkd\  &
  42.52$\pm$0.76 &
  22.26$\pm$0.31 &
  31.89$\pm$0.45 &
  34.68$\pm$1.87 &
  41.69$\pm$1.18 &
  18.09$\pm$0.88 &
  17.64$\pm$0.93 &
  9.67$\pm$0.26 \\
% \textit{+\ta\ } &
%   44.09$\pm$0.97 &
%   \textbf{19.41$\pm$0.60} &
%   35.99$\pm$0.79 &
%   \textbf{23.32$\pm$1.79} &
%   44.05$\pm$1.12 &
%   \textbf{12.97$\pm$0.43} &
%   19.37$\pm$1.73 &
%   \textbf{8.31$\pm$0.68} \\
& \textit{+ours} &
  \textbf{45.25$\pm$1.02} &
  \textbf{19.87$\pm$0.34 }&
  \textbf{37.11$\pm$0.64} &
  \textbf{24.87$\pm$1.04 }&
  \textbf{46.27$\pm$1.09} &
  \textbf{13.98$\pm$0.98} &
  \textbf{26.15$\pm$0.94} &
  \textbf{8.73$\pm$0.85} \\ \cmidrule{2-10}
& \mkd\  &
  39.36$\pm$0.70 &
  42.74$\pm$0.52 &
  32.89$\pm$0.42 &
  32.01$\pm$1.36 &
  41.04$\pm$0.93 &
  15.37$\pm$0.33 &
  19.14$\pm$1.36 &
  \textbf{8.76$\pm$0.35} \\
% \textit{+\ta\ } &
%   43.55$\pm$0.96 &
%   32.61$\pm$0.80 &
%   35.36$\pm$0.85 &
%   \textbf{20.96$\pm$1.31} &
%   41.67$\pm$1.35 &
%   \textbf{10.51$\pm$0.36} &
%   20.99$\pm$1.53 &
%   \textbf{8.20$\pm$0.98} \\
& \textit{+ours} &
  \textbf{44.85$\pm$0.80} &
  \textbf{30.08$\pm$0.57} &
  \textbf{36.79$\pm$0.70} &
  \textbf{21.84$\pm$0.33} &
  \textbf{44.19$\pm$1.17} &
  \textbf{12.56$\pm$0.53} &
  \textbf{26.10$\pm$0.75} &
  9.17$\pm$0.20 \\ \cmidrule{2-10}
& \tkd\  &
  43.74$\pm$0.84 &
  23.65$\pm$0.79 &
  34.58$\pm$0.34 &
  21.13$\pm$1.17 &
  40.44$\pm$1.40 &
  12.20$\pm$0.46 &
  14.64$\pm$0.33 &
  \textbf{6.02$\pm$0.54} \\
% \textit{+\ta\ } &
%   45.29$\pm$1.02 &
%   \textbf{19.42$\pm$0.85} &
%   34.62$\pm$0.92 &
%   \textbf{14.72$\pm$1.28} &
%   41.68$\pm$1.03 &
%   \textbf{9.29$\pm$0.75} &
%   16.66$\pm$1.66 &
%   6.88$\pm$0.36 \\
& \textit{+ours} &
  \textbf{46.21$\pm$0.86} &
  \textbf{20.45$\pm$0.57} &
  \textbf{36.26$\pm$0.71} &
  \textbf{17.01$\pm$0.89} &
  \textbf{44.22$\pm$1.08} &
  \textbf{11.79$\pm$0.57} &
  \textbf{22.00$\pm$0.97} &
  9.36$\pm$0.68 \\ \cmidrule{2-10}
 & ANCL \ & 43.15$\pm$0.49 & 32.78$\pm$1.52 & 34.32$\pm$0.41 & 36.74$\pm$1.38 & 45.16$\pm$0.32 & 20.21$\pm$0.30  & 21.84$\pm$1.33 & 11.79$\pm$0.51 \\
 &  \textit{+ours} & \textbf{46.73$\pm$0.20}  & \textbf{26.86$\pm$1.17} & \textbf{38.48$\pm$0.94} & \textbf{27.99$\pm$1.20}  & \textbf{48.25$\pm$0.33} & \textbf{16.11$\pm$0.19} & \textbf{29.67$\pm$1.13} & \textbf{10.98$\pm$0.56}
  \\ \midrule

% TINY IMAGENET

\parbox[t]{2.5mm}{\multirow{8}{*}{\rotatebox[origin=c]{90}{b) TinyImageNet200}}}  & GKD &
  32.12$\pm$0.57 &
  21.21$\pm$1.31 &
  25.75$\pm$0.65 &
  27.55$\pm$1.81 &
  34.32$\pm$1.96 &
  11.41$\pm$2.80 &
  23.30$\pm$1.59 &
  12.63$\pm$2.21 \\
% \textit{+TA} &
%   33.12$\pm$0.53 &
%   \textbf{17.50$\pm$1.76} &
%   27.16$\pm$0.87 &
%   22.62$\pm$2.03 &
%   37.01$\pm$1.84 &
%   \textbf{10.05$\pm$0.87} &
%   26.88$\pm$1.62 &
%   \textbf{11.08$\pm$0.47} \\
& \textit{+ours} &
  \textbf{33.90$\pm$0.78} &
  \textbf{17.70$\pm$1.48 }&
  \textbf{27.85$\pm$0.90} &
  \textbf{21.65$\pm$1.43} &
  \textbf{37.50$\pm$1.84} &
  \textbf{10.14$\pm$1.10} &
  \textbf{28.82$\pm$1.95} &
  \textbf{11.44$\pm$0.61} \\ \cmidrule{2-10}
& MKD &
  31.04$\pm$1.00 &
  16.90$\pm$0.75 &
  25.22$\pm$0.88 &
  25.80$\pm$2.46 &
  \textbf{33.75$\pm$2.11} &
  10.30$\pm$1.63 &
  23.42$\pm$1.95 &
  10.41$\pm$1.97 \\
% \textit{+TA} &
%   31.78$\pm$0.83 &
%   \textbf{11.42$\pm$1.23} &
%   27.05$\pm$1.11 &
%   17.55$\pm$2.36 &
%   32.23$\pm$2.28 &
%   \textbf{5.91$\pm$0.66} &
%   23.29$\pm$1.61 &
%   \textbf{7.61$\pm$1.18} \\
& \textit{+ours} &
  \textbf{32.22$\pm$0.95} &
  \textbf{11.94$\pm$1.07} &
  \textbf{27.39$\pm$1.33} &
  \textbf{16.74$\pm$1.52} &
  32.99$\pm$1.98 &
 \textbf{ 6.42$\pm$0.80} &
  \textbf{25.75$\pm$1.74} &
  \textbf{8.35$\pm$0.82} \\ \cmidrule{2-10}
& TKD &
  33.15$\pm$0.45 &
  21.05$\pm$0.39 &
  27.29$\pm$0.76 &
  23.20$\pm$2.46 &
  37.31$\pm$1.36 &
  11.83$\pm$1.31 &
  23.94$\pm$2.14 &
  10.20$\pm$1.09 \\
% \textit{+TA} &
%   33.94$\pm$0.70 &
%   17.32$\pm$0.97 &
%   28.51$\pm$0.96 &
%   17.38$\pm$2.44 &
%   37.97$\pm$1.47 &
%   \textbf{9.02$\pm$0.65} &
%   26.56$\pm$1.62 &
%   \textbf{8.05$\pm$0.36} \\
& \textit{+ours} &
  \textbf{34.58$\pm$0.96} &
  \textbf{17.27$\pm$0.54} &
  \textbf{28.71$\pm$1.06} &
  \textbf{17.37$\pm$1.78} &
  \textbf{38.41$\pm$1.52} &
  \textbf{9.16$\pm$0.62} &
  \textbf{28.10$\pm$1.97} &
  \textbf{9.56$\pm$0.65} \\ \cmidrule{2-10}
& ANCL &
  32.84$\pm$0.78 &
  27.24$\pm$1.47 &
  26.98$\pm$0.60 &
  32.45$\pm$1.18 &
  37.74$\pm$0.60 &
  17.28$\pm$2.59 &
  27.95$\pm$1.98 &
  20.91$\pm$0.92 \\
% \textit{+TA} &
%   33.85$\pm$0.55 &
%   24.02$\pm$1.39 &
%   28.42$\pm$0.48 &
%   27.26$\pm$0.98 &
%   39.50$\pm$1.18 &
%   \textbf{14.88$\pm$1.41} &
%   31.60$\pm$1.89 &
%   18.10$\pm$0.55 \\
& \textit{+ours} &
  \textbf{34.59$\pm$0.75} &
  \textbf{23.64$\pm$0.76} &
  \textbf{29.18$\pm$0.71} &
  \textbf{26.50$\pm$1.34} &
  \textbf{40.10$\pm$1.39} &
  \textbf{15.12$\pm$1.01} &
  \textbf{32.60$\pm$1.45} &
  \textbf{17.94$\pm$0.39} \\ 
  \midrule

% IMAGENET

\parbox[t]{2.5mm}{\multirow{8}{*}{\rotatebox[origin=c]{90}{c) ImageNet100}}} 
 & GKD &
  54.62$\pm$0.52 &
  25.95$\pm$0.11 &
  42.82$\pm$0.58 &
  35.39$\pm$0.88 &
  \textbf{57.94$\pm$0.90} &
  \textbf{14.47$\pm$0.83} &
  21.91$\pm$0.06 &
  \textbf{9.29$\pm$0.69} \\
& \textit{+ours} &
  \textbf{55.82$\pm$0.61} &
  \textbf{20.52$\pm$0.24} &
  \textbf{45.88$\pm$0.79} &
  \textbf{23.25$\pm$0.62} &
  57.18$\pm$0.45 &
  17.24$\pm$0.39 &
  \textbf{22.31$\pm$0.64} &
  11.28$\pm$0.98 \\ \cmidrule{2-10}
& MKD &
  54.01$\pm$0.01 &
  28.19$\pm$0.61 &
  43.39$\pm$0.66 &
  34.25$\pm$0.81 &
  \textbf{56.18$\pm$0.90} &
  14.94$\pm$0.17 &
  \textbf{26.07$\pm$0.29} &
  16.00$\pm$0.06 \\
& \textit{+ours} &
  \textbf{56.02$\pm$0.20} &
  \textbf{18.60$\pm$0.76} &
  \textbf{46.18$\pm$0.54} &
  \textbf{19.14$\pm$0.79} &
  52.05$\pm$0.24 &
  \textbf{14.23$\pm$0.46} &
  22.25$\pm$0.13 &
  \textbf{11.94$\pm$1.23} \\ \cmidrule{2-10}
& TKD &
  55.70$\pm$0.49 &
  23.55$\pm$0.35 &
  44.75$\pm$0.28 &
  32.16$\pm$0.14 &
  \textbf{54.72$\pm$0.86} &
  \textbf{10.16$\pm$0.34} &
  19.32$\pm$0.23 &
  \textbf{9.67$\pm$0.61} \\
& \textit{+ours} &
  \textbf{56.23$\pm$0.70} &
  \textbf{18.09$\pm$0.26} &
  \textbf{46.45$\pm$0.42} &
  \textbf{19.55$\pm$0.30} &
  53.85$\pm$0.39 &
  13.15$\pm$0.16 &
  \textbf{22.55$\pm$0.83} &
  9.96$\pm$0.28 \\ \cmidrule{2-10}
& ANCL &
  55.81$\pm$0.41 &
  27.13$\pm$0.50 &
  44.94$\pm$0.63 &
  34.24$\pm$1.14 &
  \textbf{60.97$\pm$0.62} &
  \textbf{15.72$\pm$0.14} &
  29.19$\pm$0.76 &
  16.12$\pm$0.27 \\
& \textit{+ours} &
  \textbf{57.41$\pm$0.22} &
  \textbf{21.54$\pm$0.50} &
  \textbf{47.49$\pm$0.35} &
  \textbf{22.61$\pm$0.40} &
  59.22$\pm$0.26 &
  17.34$\pm$0.53 &
  \textbf{29.64$\pm$0.50} &
  \textbf{14.67$\pm$0.84} \\
  \bottomrule
\end{tabular}%
}
\caption{Standard Knowledge Distillation (KD) techniques with and without our Teacher Adaptation (TA) method on different splits of a)~CIFAR100, b)~TinyImageNet200 and c)~ImageNet100. TA is generally beneficial to the CIL process, and the improvements occur most consistently in scenarios with a long sequence of equally sized tasks, where the initial model learns a weaker feature extractor.}
\label{tab:benchmarks}
\end{table*}

\section{Experiments}

\subsection{Experimental setup}
\label{sec:exp:setup}
We conduct experiments on common continual learning benchmarks, such as CIFAR100~\cite{Krizhevsky2009LearningML}, TinyImageNet200~\cite{abai2019densenet} and ImageNet-Subset~\cite{deng2009_imagenet}. We measure models’ adaptability to large shifts in data distributions on DomainNet~\cite{peng2019moment} dataset. Additionally, following FACIL~\cite{masana2022class} we construct fine-grained classification benchmark using Oxford Flowers~\cite{nilsback2008automated}, MIT Indoor Scenes~\cite{quattoni2009recognizing}, CUB-200-2011 Birds~\cite{wah2011caltech}, Stanford Cars~\cite{krause20133d}, FGVC Aircraft~\cite{maji2013fine}, and Stanford Actions~\cite{yao2011human}. Finally, we also introduce a corrupted CIFAR100 setting where data in every other task contains noise of varying severity, which allows us to measure the impact of \ta\ under varying and controllable degrees of data shift.

We create CIL scenarios by splitting the classes in each dataset into disjoint tasks. We experiment with two particular types of settings: the first type of setting is built by splitting the classes in the dataset into tasks containing an equal number of classes, while the other simulates pretaining the network and uses half of the classes as a larger first task, with subsequent tasks composed of the evenly split remaining classes. 

For all experiments, we use FACIL framework provided by Masana et al.~\cite{masana2022class}. For experiments on CIFAR100, we keep the class order from iCaRL~\cite{rebuffi2017icarl} and we use ResNet32~\cite{he2016deep}. For TinyImageNet, following~\cite{kim2023achieving}, we rescale images to 32x32 pixels and also use ResNet32. For the other datasets, we use ResNet18~\cite{he2016deep}. We always use the same hyperparameters for all variants within the single KD method unless stated otherwise, we report the results averaged over three runs with different random seeds.

In every setup, we train the network on each new task for 200 epochs with batch size 128. We use SGD optimizer without momentum or weight decay, with a learning rate scheduler proposed by Zhou et al.~\cite{zhou2023pycil}, where the initial learning rate of 0.1 is decreased 10x after 60th, 120th and 160th epoch. For experiments conducted on CIFAR100 and TinyImageNet200 in \Cref{tab:benchmarks} we also employ a warmup phase~\cite{Kumar2022finetunedistort} for the new classification head. In the Appendix, we provide the ablation study of the warmup with different benchmarks, alongside the details of its implementation and discussion on the method. Additionally, we provide the evaluation of our method with different model architectures and batch sizes.

\noindent\textbf{Evaluation metrics.}
The average incremental accuracy at task $k$ is defined as $A_{k}\!=\!\frac{1}{k} \sum_{j=1}^{k}a_{k,j}$, where
${a_{k,j}} \in [0, 1]$ be the accuracy of the $j$-th task ($j \leq k$) after training the network sequentially for $k$ tasks~\cite{aljundi2017expert}. Overall average incremental accuracy $Acc_{Inc}$ is the mean value from all tasks. We also report \textit{average forgetting} as defined in~\cite{chaudhry2018riemannian}, while the $Forg_{Inc}$ is similarly the mean value from all tasks. We provide results with additional metrics such as final accuracy $Acc_{Final}$ and final forgetting $Forg_{Final}$ in the Appendix.

\begin{table*}[!th]
\small
\centering

%\resizebox{\textwidth}{!}{%
\begin{tabularx}{\textwidth}{@{}X *6{>{\centering\arraybackslash}X}@{}}
\toprule
 &
  \multicolumn{2}{c}{6 tasks, 20 classes each} &
  \multicolumn{2}{c}{12 tasks, 10 classes each} &
  \multicolumn{2}{c}{24 tasks, 5 classes each} \\ \cmidrule{2-7}
 &
  \multicolumn{1}{c}{$Acc_{Inc} \uparrow$} &
  \multicolumn{1}{c}{$Forg_{Inc} \downarrow$}  &
  \multicolumn{1}{c}{$Acc_{Inc} \uparrow$} &
  \multicolumn{1}{c}{$Forg_{Inc} \downarrow$} &
  \multicolumn{1}{c}{$Acc_{Inc} \uparrow$} &
  \multicolumn{1}{c}{$Forg_{Inc} \downarrow$} \\ \midrule
\gkd\  &
  56.03$\pm$3.75 &
  \textbf{30.23$\pm$0.72} &
  44.03$\pm$1.53 &
  42.70$\pm$2.32 &
  30.14$\pm$4.00 &
  51.95$\pm$2.28 \\
\textit{+ours} &
  \textbf{57.24$\pm$1.79} &
  33.12$\pm$1.76 &
  \textbf{50.80$\pm$1.27} &
  \textbf{37.58$\pm$3.01} &
  \textbf{42.86$\pm$1.82} &
  \textbf{39.01$\pm$1.55} \\
% \textit{+\ta\ +\wu\ } &
%   52.96$\pm$1.20 &
%   38.77$\pm$2.45 &
%   49.46$\pm$2.09 &
%   \textbf{35.81$\pm$2.02} &
%   \textbf{45.17$\pm$0.65} &
%   \textbf{33.27$\pm$1.89} \\ 
  \midrule
MKD &
  \textbf{60.12$\pm$1.59} &
  \textbf{26.42$\pm$1.60} &
  45.19$\pm$2.91 &
  43.81$\pm$1.56 &
  32.61$\pm$0.55 &
  55.91$\pm$1.77 \\
\textit{+ours} &
  55.77$\pm$2.23 &
  31.43$\pm$2.28 &
  \textbf{51.16$\pm$1.84} &
  \textbf{34.89$\pm$2.60} &
  \textbf{45.49$\pm$1.82} &
  \textbf{34.48$\pm$0.89} \\
% \textit{+TA+WU} &
%   56.41$\pm$1.04 &
%   34.47$\pm$0.71 &
%   50.40$\pm$3.57 &
%   35.74$\pm$3.45 &
%   \textbf{45.80$\pm$2.48} &
%   \textbf{32.48$\pm$2.15} \\ 
  \midrule
\tkd\  &
  56.69$\pm$3.36 &
  33.86$\pm$1.36 &
  46.28$\pm$1.42 &
  43.38$\pm$0.82 &
  32.17$\pm$2.71 &
  51.72$\pm$1.95 \\
\textit{+ours} &
  \textbf{57.99$\pm$1.88} &
  \textbf{33.79$\pm$1.80} &
  \textbf{51.66$\pm$1.63} &
  \textbf{35.43$\pm$2.75} &
  \textbf{43.95$\pm$2.60} &
  \textbf{33.28$\pm$1.86}
% \textit{+\ta\ +\wu\ } &
%   53.68$\pm$1.56 &
%   39.73$\pm$2.92 &
%   50.12$\pm$2.31 &
%   \textbf{34.64$\pm$1.99} &
%   37.22$\pm$6.11 &
%   \textbf{24.53$\pm$3.01} 
\\
\bottomrule
\end{tabularx}%
%}
\caption{Average task-agnostic accuracy and forgetting for KD-based CL methods on fine-grained classification datasets.}
\label{tab:finegrained_main}
\end{table*}

\begin{table*}[!th]
\centering

\resizebox{\textwidth}{!}{%
\begin{tabular}{lrrrrrrrr}
\toprule
 &
  \multicolumn{2}{c}{6 tasks} &
  \multicolumn{2}{c}{6 tasks, pretrained} &
  \multicolumn{2}{c}{12 tasks} &
  \multicolumn{2}{c}{12 tasks, pretrained} \\ \cmidrule{2-9}
 &
  \multicolumn{1}{c}{$Acc_{Inc} \uparrow$} &
  \multicolumn{1}{c}{$Forg_{Inc} \downarrow$} &
  \multicolumn{1}{c}{$Acc_{Inc} \uparrow$} &
  \multicolumn{1}{c}{$Forg_{Inc} \downarrow$} &
  \multicolumn{1}{c}{$Acc_{Inc} \uparrow$} &
  \multicolumn{1}{c}{$Forg_{Inc} \downarrow$} &
  \multicolumn{1}{c}{$Acc_{Inc} \uparrow$} &
  \multicolumn{1}{c}{$Forg_{Inc} \downarrow$}  \\ \midrule
  \gkd\  &
  18.63$\pm$0.27 &
  \textbf{23.27$\pm$0.26} &
  43.27$\pm$0.10 &
  36.83$\pm$0.88 &
  14.45$\pm$0.25 &
  \textbf{29.04$\pm$0.37} &
  35.98$\pm$0.96 &
  43.00$\pm$1.66 
  \\
\textit{+ours } &
  \textbf{19.55$\pm$0.42} &
  27.22$\pm$0.21 &
  \textbf{43.52$\pm$0.17} &
  \textbf{34.98$\pm$0.19} &
  \textbf{16.25$\pm$0.46} &
  33.03$\pm$0.19 &
  \textbf{38.89$\pm$0.52} &
  \textbf{41.10$\pm$0.42}
  \\
% \textit{+\ta\ +\wu\ } &
%   18.68$\pm$0.36 &
%   31.12$\pm$0.38 &
%   39.00$\pm$0.12 &
%   49.44$\pm$0.37 &
%   16.22$\pm$0.45 &
%   32.63$\pm$0.11 &
%   34.24$\pm$0.42 &
%   44.17$\pm$0.37
%   \\
  \midrule

  \tkd\  &
  19.12$\pm$0.26 &
  \textbf{26.66$\pm$0.38} &
  42.42$\pm$0.10 &
  \textbf{40.83$\pm$1.11} &
  16.31$\pm$0.55 &
  37.32$\pm$0.87 &
  38.15$\pm$0.22 &
  42.28$\pm$0.50
  \\
\textit{+ours } &
  \textbf{19.57$\pm$0.10} &
  29.96$\pm$0.05 &
  \textbf{42.75$\pm$0.14} &
  41.79$\pm$0.65 &
  \textbf{16.74$\pm$0.55} &
  \textbf{33.32$\pm$0.47} &
  \textbf{39.06$\pm$0.33} &
  \textbf{40.19$\pm$0.85}
  \\
% \textit{+\ta\ +\wu\ } &
%   18.65$\pm$0.27 &
%   32.33$\pm$0.49 &
%   40.62$\pm$0.07 &
%   46.86$\pm$0.39 &
%   16.40$\pm$0.56 &
%   \textbf{32.87$\pm$0.34} &
%   36.81$\pm$0.97 &
%   41.98$\pm$0.95
%   \\

  \midrule

  %MKD  & 
  %\\
  %\textit{+\ta\ } &

  %\textit{+\ta\ + \wu\} &
MKD             
  & \textbf{18.74$\pm$0.52} 
  & \textbf{19.10$\pm$0.40}  
  & \textbf{45.70$\pm$0.30}    
  & \textbf{27.48$\pm$0.60}  
  & 13.45$\pm$0.53 
  & \textbf{27.23$\pm$0.98} 
  & \textbf{39.14$\pm$0.21} 
  & 36.53$\pm$1.00  \\
  \textit{+ours } 
  & 18.04$\pm$0.16 
  & 22.70$\pm$0.25 
  & 42.91$\pm$0.04 
  & 29.43$\pm$0.04 
  &\textbf{15.30$\pm$0.35}  
  & 28.71$\pm$0.26 
  & 37.84$\pm$0.35 
  & \textbf{34.09$\pm$0.39} \\
  % \textit{+\ta\ + \wu\ } & 19.09$\pm$0.76  & 27.46$\pm$0.33 & 42.21$\pm$0.92  & 38.07$\pm$0.54  & 15.82$\pm$0.31  & 30.23$\pm$0.40  &               &      \\
  
  \midrule
  ANCL  & 19.58$\pm$0.46 & \textbf{25.63$\pm$0.20}  & \textbf{42.90$\pm$0.84}  & \textbf{37.28$\pm$1.86} & 14.82$\pm$0.41 & \textbf{33.46$\pm$0.40} & 33.34$\pm$0.55 & 49.05$\pm$0.65 
  \\
  \textit{+ours } & \textbf{20.34$\pm$0.40}  & 30.73$\pm$0.44 & 42.67$\pm$0.51 & 38.56$\pm$1.24 & \textbf{17.19$\pm$0.06} & 37.40$\pm$0.38 & \textbf{35.81$\pm$0.18} & \textbf{43.84$\pm$0.23}
  \\
  % \textit{+\ta\ +\wu\ } & 20.39$\pm$0.09 & 32.83$\pm$0.66 & 42.44$\pm$0.80 & 37.09$\pm$1.33 &  &  &  & 
  % \\
  
 \bottomrule
\end{tabular}%
}
\caption{Average task-agnostic accuracy and forgetting for KD and TA under significant semantic drift on DomainNet. We test scenarios with 6 tasks of 50 classes and 12 tasks of 25 classes, both when training from scratch and starting from pretrained model. Aside from MKD, TA generally leads to better results.}
\label{tab:domainnet_main}
\end{table*}
\subsection{Standard CIL benchmarks}
\label{sec:exp:benchmarks}
We evaluate knowledge distillation approaches described in \Cref{sec:kd_in_cl} on the standard CIL benchmarks CIFAR100, TinyImageNet200 and ImageNet100, using different class splits. We present the results in \Cref{tab:benchmarks} a), b) and c) respectively. We also provide results for more settings and ablation study of our method on those datasets in the Appendix.

In most settings, we observe that our method improves upon the baseline knowledge distillation. We notice that the improvement TA is generally more significant in settings with a larger number of tasks and an equal split of the classes. In settings with half the classes presented in the first task, the gains from TA are sometimes not that visible, as in this case the initial model already learns a good feature extractor, and the distribution of its normalization statistics after the first task is a better approximation of the statistics for the whole dataset. TA sometimes underperforms with MKD, which might be caused by the fact that the loss formula of MKD uses sigmoid function, and the differences between the probabilities for KD loss are insignificant if the values of logits are not small and centered around zero, which is not guaranteed without imposing additional learning constraints.

\subsection{TA under severe distribution shifts} %Severe shift of distribution
\label{sec:exp:shift}

Motivated by the continual learning settings in which the data distribution changes significantly across the tasks, we conduct a series of experiments to empirically verify the benefits of our method.

\subsubsection{Fine-grained classification datasets}
\label{sec:exp:finegrained}

We evaluate TA on fine-grained classification tasks using six datasets: Stanford Actions, FGVC Aircraft, Stanford Cars, CUB-200-2011 Birds, MIT Indoor Scenes and Oxford Flowers. We create CIL tasks by randomly sampling a subset of classes from each dataset, in the above-mentioned order. We sample the classes without replacement, and to obtain the settings with 12 or 24 tasks we repeat the procedure. For this set of experiments, we start from ResNet18 checkpoint pretrained on ImageNet.

We conduct experiments using splits of 24 tasks with 5 classes each, 12 tasks with 10 classes each, and 6 tasks with 20 classes each. We show the results in~\Cref{tab:finegrained_main}. Consistently with the results from \Cref{sec:exp:benchmarks}, our method generally improves upon the base KD, with the improvements being more visible on the longer tasks. 

We omit ANCL method from our analysis, as we were unable to obtain sufficiently good results with its official implementation. We provide the results of additional experiments conducted with reverse order of datasets and with the full datasets used as a single task in the Appendix. 

\subsubsection{Large domain shifts with DomainNet}
\label{sec:exp:domainnet}

\begin{figure*}[!ht]
    \centering
    \begin{subfigure}[b]{0.33\textwidth}
        \centering
        \includegraphics[width=\textwidth]{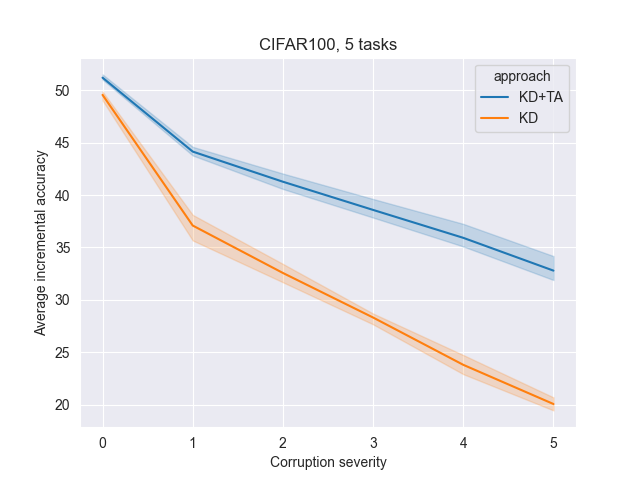}
          \end{subfigure}
    \begin{subfigure}[b]{0.33\textwidth}
        \centering
        \includegraphics[width=\textwidth]{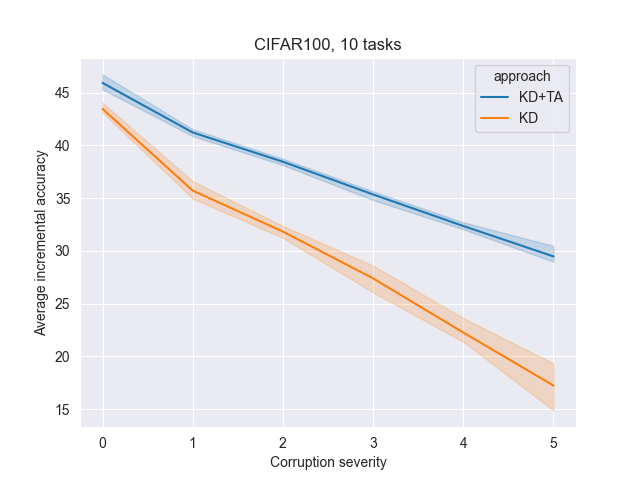}
    \end{subfigure}
    \begin{subfigure}[b]{0.33\textwidth}
        \centering
        \includegraphics[width=\textwidth]{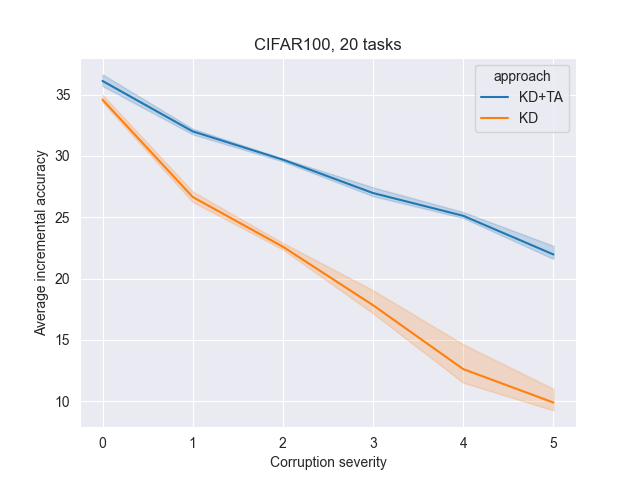}
    \end{subfigure}

   \caption{
   Average incremental accuracy for standard KD and our method of \ta\ under varying strength of data shift on splits of CIFAR100. As the noise strengthens, the gap between \ta\ and standard KD widens, indicating that our method leads to more robust learning in case of data shifts. We obtain data shifts by adding noise of varying strength to every other task, using the Gaussian noise and noise severity scale proposed by Michaelis et al.~\cite{michaelis2019dragon}. 
   }
   \label{fig:corruptions}
\end{figure*}
To verify the effectiveness of teacher adaptation for continual learners under significant data distribution shifts, we use DomainNet~\cite{peng2019moment} as our evaluation dataset. DomainNet consists of images from 6 domains and 345 classes. We select the first 50 classes and create each task from a different domain, resulting in more severe data drift between tasks in CIL. This allows us to measure how well the models can adapt to new data distributions. We use ResNet18 and compare two settings: training from scratch and from starting from the model pretrained on ImageNet. ~\Cref{tab:domainnet_main} shows the results of our experiments. Consistently with the results from previous Sections, we find that, aside from MKD, TA generally performs better than the baselines, and the differences are more visible when training on 12 tasks, where the model is exposed to more changes in the data distribution.

\subsubsection{Varying the strength of the distribution shift}
\label{sec:exp:corruptions}

We create CIL settings with controllable levels of data distribution shift between subsequent tasks by corrupting every other task. We split CIFAR100 into 5, 10, and 20 tasks of equal size and add Gaussian noise to every other task, so that in subsequent tasks the data distribution changes from clean to noisy or vice versa. We obtain varying strength of distribution shift by using different levels of noise severity, following the methodology of Michaelis et al.~\cite{michaelis2019dragon}.

We show the results of this experiment in \Cref{fig:corruptions}. We see that as the noise severity increases, the gap between standard KD and \ta\ widens, indicating that our method is better suited to more challenging scenarios of learning under extreme data distribution shifts.

\subsection{Detailed analysis}
\label{sec:exp:bn_ablations}

\subsubsection{Alternatives to batch normalization} We conduct a series of ablation experiments on CIFAR100 split into 10 tasks to justify the validity of our method over other potential solutions for adaptation of batch normalization layers. The results of those experiments are shown in~\Cref{tab:bn_ablations}. We compare the following settings: 1)~standard training with batch normalization statistics from the previous task fixed in the teacher model, but updated in the student model,  2)~batch normalization layers removed, 3)~batch normalization statistics fixed in both models after learning the first task, 4)~batch normalization layers replaced with LayerNorm~\cite{ba2016layer} or 5)~GroupNorm~\cite{wu2018group} layers, and finally 6)~our solution of Teacher Adaptation. 

\begin{table}[!h]
\centering
\resizebox{\columnwidth}{!}{%
\begin{tabular}{@{}lrrrr@{}}
\toprule
$clip=100$       & \multicolumn{2}{c}{$\lambda=5$}                                     & \multicolumn{2}{c}{$\lambda=10$}                                    \\ \midrule
                 & \multicolumn{1}{c}{$Acc_{Final} \uparrow$} & \multicolumn{1}{c}{$Acc_{Inc} \uparrow$} & \multicolumn{1}{c}{$Acc_{Final} \uparrow$} & \multicolumn{1}{c}{$Acc_{Inc} \uparrow$} \\ \cmidrule{2-5}
1) \textbf{\gkd\ }     & 25.47$\pm$0.57                    & 41.59$\pm$0.32                  & 27.96$\pm$0.34                    & 42.28$\pm$0.67                  \\
2) \textbf{-BN}     & 0.33$\pm$1.15                     & 2.01$\pm$2.67                   & 0.33$\pm$1.15                     & 2.85$\pm$3.81                   \\
3) \textbf{fix BN}  & -                                 & -                               & -                                 & -                               \\
4) \textbf{-BN +LN} & 21.94$\pm$0.95                    & 34.7$\pm$0.48                   & 22.76$\pm$1.05                    & 34.48$\pm$0.15                  \\
5) \textbf{-BN +GN} & 21.92$\pm$0.46 & 32.15$\pm$0.16 & 22.01$\pm$0.82 & 31.71$\pm$0.35 \\
6) \textbf{+\ta\ }    & \textbf{31.39$\pm$0.17}           & \textbf{44.98$\pm$0.38}         & \textbf{31.85$\pm$0.10}           & \textbf{44.06$\pm$0.69}         \\ \midrule
$clip=1$         & \multicolumn{2}{c}{$\lambda=5$}                                     & \multicolumn{2}{c}{$\lambda=10$}                                    \\ \midrule
                 & \multicolumn{1}{c}{$Acc_{Final} \uparrow$} & \multicolumn{1}{c}{$Acc_{Inc} \uparrow$} & \multicolumn{1}{c}{$Acc_{Final}  \uparrow$} & \multicolumn{1}{c}{$Acc_{Inc} \uparrow$} \\ \cmidrule{2-5}
1) \textbf{\gkd\ }     & 20.80$\pm$0.56                    & 34.28$\pm$0.24                  & 27.96$\pm$0.34                    & 42.28$\pm$0.67                  \\
2) \textbf{-BN}     & 19.47$\pm$0.18                    & 29.83$\pm$0.53                  & 0.33$\pm$1.15                     & 2.85$\pm$3.81                   \\
3) \textbf{fix BN}  & 20.21$\pm$0.31                    & 32.07$\pm$0.20                  & -                                 & -                               \\
4) \textbf{-BN +LN} & 18.49$\pm$1.41                    & 30.39$\pm$0.72                  & 22.76$\pm$1.05                    & 34.48$\pm$0.15                  \\
5) \textbf{-BN +GN} & 16.17$\pm$0.89 & 32.15$\pm$0.16 & 15.73$\pm$1.01 & 25.07$\pm$1.10  \\  
6) \textbf{+\ta\ }    & \textbf{24.19$\pm$0.90}           & \textbf{36.13$\pm$0.24}         & \textbf{31.85$\pm$0.10}           & \textbf{44.06$\pm$0.69}         \\ \bottomrule
\end{tabular}%
}
\caption{Results for different solutions to the problem of diverging batch normalization layers when using knowledge distillation in continual learning. We use \gkd\  with different $\lambda$ and gradient clipping values. We compare the baseline with variants without batch normalization layers, with batch normalization statistics fixed after the first task and with batch normalization layers replaced with LayerNorm or GroupNorm. "-" indicates that training crashes due to instability. \ta\ is the only solution that improves upon the baseline.}
\label{tab:bn_ablations}
\end{table}

Fixing or removing BatchNorms leads to unstable training. This can be partially fixed by setting a high gradient clipping value or lowering the lambda parameter, but both solutions lead to much worse network performance. Different normalization layers enable stable training, but ultimately converge to much worse solutions than the network with BatchNorm.
Our solution is the only one that improves over different values of $\lambda$ and does not require controlling the gradients by clipping the high values.

\subsubsection{Alternative methods of teacher adaptation.} 
We study alternative methods of adapting the teacher model and try pretraining ($P$) or continuously training ($CT$) the teacher model. For pertaining, we train the teacher on the new data in isolation for a few epochs before the training of the main model. During continuous training, we update the teacher alongside the main model using the same batches of new data. With both approaches, we set a lower learning rate for the teacher. We conduct those experiments by training either the full teacher model ($FM$) or only its batch normalization layers ($BN$). Finally, to isolate the impact of changing batch normalization statistics and training model parameters, we repeat all the experiments with fixed batch normalization statistics ($fix\ BN$). 

We conduct our experiment on CIFAR100 split into 10 tasks and present the results in \Cref{tab:ta_ablations}. Alternative solutions perform within the standard deviation of TA with tuned hyperparameters, but the values of the hyperparameters for those models (described in the Appendix) are also very small, indicating that the teacher model doesn't change much. Upon closer study, we find that it's mostly batch normalization statistics that change throughout the training. Therefore, knowing that other successful methods from test-time adaptation~\cite{TENT} use a similar approach, we continue with Teacher Adaptation based on batch normalization layers, as it does not require any hyperparameter tuning or additional pretraining epochs.

\begin{table}[!h]
\centering
\resizebox{\columnwidth}{!}{%
\begin{tabular}{@{}lrrrr@{}}
\toprule
Method             & \multicolumn{1}{c}{$Acc_{Final} \uparrow$} & \multicolumn{1}{c}{$Acc_{Inc} \uparrow$} & \multicolumn{1}{c}{$Forg_{Final} \downarrow$} & \multicolumn{1}{c}{$Forg_{Inc} \downarrow$} \\ \midrule
Base                                & 27.53$\pm$0.15 & 42.22$\pm$0.38 & 31.28$\pm$1.64 & 23.11$\pm$1.58 \\ \midrule
P-FM        & 31.54$\pm$0.67 & 43.46$\pm$0.72 & 24.18$\pm$1.17 & 20.80$\pm$1.51 \\
\textit{+fix BN}        & 28.02$\pm$0.60 & 42.33$\pm$0.53 & 29.91$\pm$1.27 & 22.66$\pm$0.95 \\ \midrule
P-BN           & 31.16$\pm$0.54 & 43.64$\pm$0.77 & 24.44$\pm$0.96 & 20.13$\pm$0.75 \\
\textit{+fix BN}        & 27.62$\pm$0.48 & 42.12$\pm$0.38 & 29.95$\pm$1.64 & 22.50$\pm$0.95 \\ \midrule
CT-FM & 31.37$\pm$0.94 & 43.38$\pm$0.77 & 24.34$\pm$1.37 & 20.93$\pm$1.58 \\
\textit{+fix BN}        & 28.17$\pm$0.49 & 42.29$\pm$0.42 & 29.79$\pm$1.02 & 22.55$\pm$0.67 \\ \midrule
CT-BN    & 31.35$\pm$0.63 & 43.69$\pm$0.76 & 24.29$\pm$0.61 & 20.23$\pm$0.59 \\
\textit{+fix BN}        & 27.33$\pm$0.50 & 42.09$\pm$0.45 & 30.20$\pm$1.73 & 22.50$\pm$0.85 \\ \midrule
TA & \textbf{32.15$\pm$0.12}           & \textbf{44.31$\pm$0.26}         & \textbf{23.55$\pm$0.51}            & \textbf{19.85$\pm$0.93}          \\ \bottomrule
\end{tabular}%
}
\caption{Ablation study of different ways to adapt the teacher model. Our method achieves the best results while requiring no additional hyperparameters. We try teacher adaptation during pretraining (P) and continuous training (CT). We train either full model (FM) or only batch normalization layers (BN). \emph{fix BN} indicates fixed BN statistics.}
\label{tab:ta_ablations}
\end{table}

\section{Conclusions}
 
We propose Teacher Adaptation, a simple yet effective method to improve the performance of knowledge distillation-based methods in exemplar-free class-incremental learning. Our method continuously updates the teacher network by adjusting batch normalization statistics during learning a new task both for the currently learning model and the teacher model saved after learning the previous tasks. This mitigates the changes in the model caused by knowledge distillation loss that arise as the current learner is continuously trying to compensate for the modified normalization statistics. We further improve the stability of the model by introducing a warm-up phase at the beginning of the task, where a new classification head is trained in isolation before finetuning the whole model. The warm-up phase ensures that the initialization of the weights is not random in the initial phases of training, and reduces the gradient updates to the whole model.
We conduct experiments with Teacher Adaptation on several class-incremental benchmarks and show that it consistently improves the results for different knowledge distillation-based methods in an exemplar-free setting. Moreover, our method can be easily added to the existing class-incremental learning solutions and induces only a slight computational overhead. 

\paragraph{Discussion}
Since the introduction of Learning without Forgetting, KD-based methods have emerged as effective solutions to mitigate forgetting in CIL models. Several approaches, such as iCaRL, EEIL, BiC, LUCIR, and SSIL, have integrated KD with exemplars, which helps maintain a balanced discrepancy between the teacher and student models.
In scenarios where a sufficient number of exemplars are available, teacher adaptation may not be required, as their presence in the training data mitigates the divergence between the normalization statistics of the subsequent tasks.
Nevertheless, our research is dedicated exclusively to the exemplar-free setting, in which we investigate techniques that do not rely on storing exemplars. To the best of our knowledge, we are the first to propose the adaptation of the teacher model within the context of KD-based exemplar-free CIL.

\paragraph{Impact} 
Our method focuses on exemplar-free scenarios, and therefore we alleviate the issues with storing potentially confidential, private, or sensitive data.
However, we recognize that machine learning algorithms can be harmful if applied carelessly, and we encourage practitioners to carefully check training data and the models to ensure that the results of their work do not perpetuate biases or discriminate against any minority.

All our work was conducted using publicly available datasets and open-source code. To allow other researchers to build on our work and validate the results, we will share the code for the experiments in this paper on GitHub upon acceptance.

\section*{Acknowledgements}
Filip Szatkowski and Tomasz Trzcinski are supported by National Centre of Science (NCP, Poland) Grant No. 2022/45/B/ST6/02817. Tomasz Trzciński is also supported by NCP Grant No. 2020/39/B/ST6/01511.

{\small
\bibliographystyle{ieee_fullname}
\bibliography{egbib}
}

\clearpage

\input{Appendix}

\end{document}

%% file: Appendix.tex
\begin{appendix}

\twocolumn[\appendixhead]

This document contains additional experimental results and analysis.

\section{ANCL implementation with TA}
We follow the provided implementation of ANCL~\cite{kim2023achieving} based on the FACIL~\cite{masana2022class} framework. We only apply the adaptation to the main teacher network fitted on the previous data, and, in cases where we use warmup, we only apply it to the new head of the main network. The auxiliary network does not undergo the teacher adaptation process, as it is already fit for the current task, and we do not see a reason to apply the warmup to this network. We keep the same training schedule and hyperparameters as described in \Cref{sec:exp:setup}.

\section{Warmup phase}
As mentioned in \Cref{sec:exp:setup}, we experimented with applying the warmup phase, a technique that recently appeared in several works~\cite{Kumar2022finetunedistort,berariu2021study,sun2023improving}. With this method, before training the full student model on the new data, we first finetune the classification head added for this new task in isolation, while the rest of the network is kept frozen. 
Compared to the standard practice of finetuning the full student model on the new task, starting with a warmup phase reduces the initial cross-entropy loss at the start of the new task, and avoids overwriting the knowledge from the previous tasks with large gradient updates caused by random classification head initialization.

For training the new head during the warmup phase, we use SGD optimizer and OneCycle scheduler~\cite{smith2019super} with cosine annealing, maximum learning rate of 0.1 and the number of epochs of increasing learning rate set as 40. We train the new head for 200 epochs with early stopping and freeze updates of batch normalization statistics in the model during the training.

In \Cref{sec:app:experiments}, we evaluate different CL benchmarks with TA and warmup separately. We generally find that the warmup works complementary to our method when we use smaller networks such as ResNet32 and train on smaller datasets. We hypothesize that for the settings with a larger dataset and a larger network better initialization of the new head does not translate to better performance in CIL because the overall magnitude of updates required to train the model in such settings is greater and leads to overwriting the initialization over the course of the training.

\vspace{2cm}

\section{Ablation studies of Teacher Adaptation}

\subsection{Teacher Adaptation with different batch sizes.} Larger batch sizes should lead to better estimation of the global statistics in the batch normalization layer, which, in theory, should lead to reduced effectiveness of TA. To investigate the impact of the batch size on our method, we train the model on CIFAR100 split into 10 tasks with different batch sizes, tuning the learning rate and $\lambda$ for each batch size. We present the results in \Cref{tab:bs_ablations}. Regardless of the batch size, TA constantly improves upon standard training, and the difference between the two approaches does not scale with batch size.

\begin{table}[!h]
\centering
\resizebox{\columnwidth}{!}{%
\begin{tabular}{@{}lrrrr@{}}
\toprule
Batch size & \multicolumn{1}{c}{$Acc_{Final} \uparrow$} & \multicolumn{1}{c}{$Acc_{Inc} \uparrow$} & \multicolumn{1}{c}{$Forg_{Final} \downarrow$} & \multicolumn{1}{c}{$Forg_{Inc} \downarrow$} \\ 
\midrule
32           & 28.94$\pm$0.29 & 44.24$\pm$0.27 & 32.57$\pm$0.92 & 23.89$\pm$1.19 \\
\textit{+TA} & \textbf{32.58$\pm$0.80} & \textbf{45.69$\pm$0.35} & \textbf{25.44$\pm$0.94} & \textbf{20.92$\pm$0.53} \\ 
\midrule
64           & 28.92$\pm$0.57 & 43.89$\pm$0.27 & 30.48$\pm$1.11 & 21.70$\pm$0.39 \\
\textit{+TA} & \textbf{32.15$\pm$1.24} & \textbf{44.71$\pm$0.57} & \textbf{23.43$\pm$1.97} & \textbf{19.16$\pm$0.67} \\ 
\midrule
128          & 28.27$\pm$0.44 & 42.52$\pm$0.76 & 29.59$\pm$0.92 & 22.26$\pm$0.31 \\
\textit{+TA} &\textbf{ 31.92$\pm$0.86 }& \textbf{44.09$\pm$0.97} & \textbf{22.65$\pm$1.32} & \textbf{19.41$\pm$0.60} \\ 
\midrule
256          & 25.13$\pm$0.46 & 40.44$\pm$0.11 & 40.06$\pm$0.38 & 31.39$\pm$0.98 \\
\textit{+TA} & \textbf{30.56$\pm$0.68} & \textbf{43.28$\pm$0.23} & \textbf{28.24$\pm$0.64} & \textbf{23.94$\pm$0.59} \\ 
\midrule
512          & 23.01$\pm$0.89 & 37.44$\pm$0.34 & 40.57$\pm$1.20 & 32.03$\pm$0.69 \\
\textit{+TA} &\textbf{ 27.52$\pm$1.03} & \textbf{39.86$\pm$0.45} & \textbf{29.31$\pm$0.86} & \textbf{25.03$\pm$0.55} \\ \bottomrule
\end{tabular}%
}
\caption{Teacher adaptation with different batch sizes. Our method improves upon the standard training regardless of the batch size.}
\label{tab:bs_ablations}
\end{table}

\subsection{Different CNN architectures with Teacher Adaptation.} 
To test the robustness of our method, in addition to experiments with ResNets, we test it with other standard convolutional neural networks. We evaluate our method with MobileNetV2~\cite{sandler2018mobilenetv2}, MobileNetV3-small~\cite{howard2019searching} and VGG11~\cite{simonyan2014very}. We use CIFAR100 and ImageNet100 split into 10 equally sized tasks and train the networks as described in \Cref{sec:exp:setup}. We show the results of those experiments in the \Cref{tab:networks_ablation}. The final results obtained with tested networks are lower than with our standard baselines, as do not tune their hyperparameters extensively, but we still notice that using TA constantly improves the results, regardless of the architecture we use.

\begin{table}[!h]
\centering
\resizebox{\columnwidth}{!}{%
\begin{tabular}{lrrrr}
\toprule
\multicolumn{5}{c}{\large{CIFAR100}}                                                                                                                               \\ \midrule
Network      & \multicolumn{1}{c}{$Acc_{Final} \uparrow$} & \multicolumn{1}{c}{$Acc_{Inc} \uparrow$} & \multicolumn{1}{c}{$Forg_{Final} \downarrow$} & \multicolumn{1}{c}{$Forg_{Inc} \downarrow$} \\ \midrule
MobilenetV2   & 12.47$\pm$0.69                    & 25.71$\pm$1.05                  & 49.86$\pm$0.61                     & 41.23$\pm$1.25                   \\
\textit{+TA}  & \textbf{20.62$\pm$1.01}           & \textbf{31.91$\pm$0.57}         & \textbf{26.66$\pm$1.34}            & \textbf{22.31$\pm$0.56}          \\ \midrule
MobilenetV3-s & 16.59$\pm$0.09                    & 29.86$\pm$0.06                  & 38.89$\pm$0.42                     & 30.99$\pm$0.12                   \\ 
\textit{+TA}  & \textbf{21.22$\pm$0.91}           & \textbf{32.24$\pm$0.60}         & \textbf{20.98$\pm$0.13}            & \textbf{16.84$\pm$1.19}          \\ \midrule
ResNet32      & 28.27$\pm$0.44                    & 42.52$\pm$0.76                  & 29.59$\pm$0.92                     & 22.26$\pm$0.31                   \\
\textit{+TA}  & \textbf{31.92$\pm$0.86}           & \textbf{44.09$\pm$0.97}         & \textbf{22.65$\pm$1.32}            & \textbf{19.41$\pm$0.60}          \\ \midrule
VGG11         & 17.10$\pm$0.12                    & 34.34$\pm$0.39                  & 62.36$\pm$0.62                     & 52.99$\pm$0.97                   \\
\textit{+TA}  & \textbf{24.87$\pm$0.35}           & \textbf{42.50$\pm$0.15}         & \textbf{43.96$\pm$0.62}            & \textbf{32.92$\pm$0.51}          \\ \midrule

\multicolumn{5}{c}{\large{ImageNet100}} \\ \midrule
Network      & \multicolumn{1}{c}{$Acc_{Final} \uparrow$} & \multicolumn{1}{c}{$Acc_{Inc} \uparrow$} & \multicolumn{1}{c}{$Forg_{Final} \downarrow$} & \multicolumn{1}{c}{$Forg_{Inc} \downarrow$} \\ \midrule
ResNet18     & 40.99$\pm$0.45                    & 54.82$\pm$0.56                  & 31.99$\pm$0.87                     & 25.90$\pm$0.65                   \\
\textit{+TA} & \textbf{42.69$\pm$0.53}           & \textbf{55.84$\pm$0.51}         & \textbf{25.26$\pm$0.23}            & \textbf{20.43$\pm$0.33}          \\ \midrule
VGG11        & 24.35$\pm$0.32                    & 44.72$\pm$0.50                  & 61.62$\pm$0.10                     & 46.73$\pm$0.31                   \\
\textit{+TA} & \textbf{30.67$\pm$0.31}           & \textbf{48.81$\pm$0.29}         & \textbf{50.04$\pm$0.76}            & \textbf{37.60$\pm$0.56}         \\ \bottomrule
\end{tabular}%

}
\caption{Teacher Adaptation with different network architectures on CIFAR100 and ImageNet100 split into 10 equally sized tasks.}
\label{tab:networks_ablation}
\end{table}

\subsection{Batch normalization statistics investigation.} 

\begin{figure*}[!ht]
    \centering
    \begin{subfigure}[b]{0.49\textwidth}
        \centering
        \includegraphics[width=\textwidth]{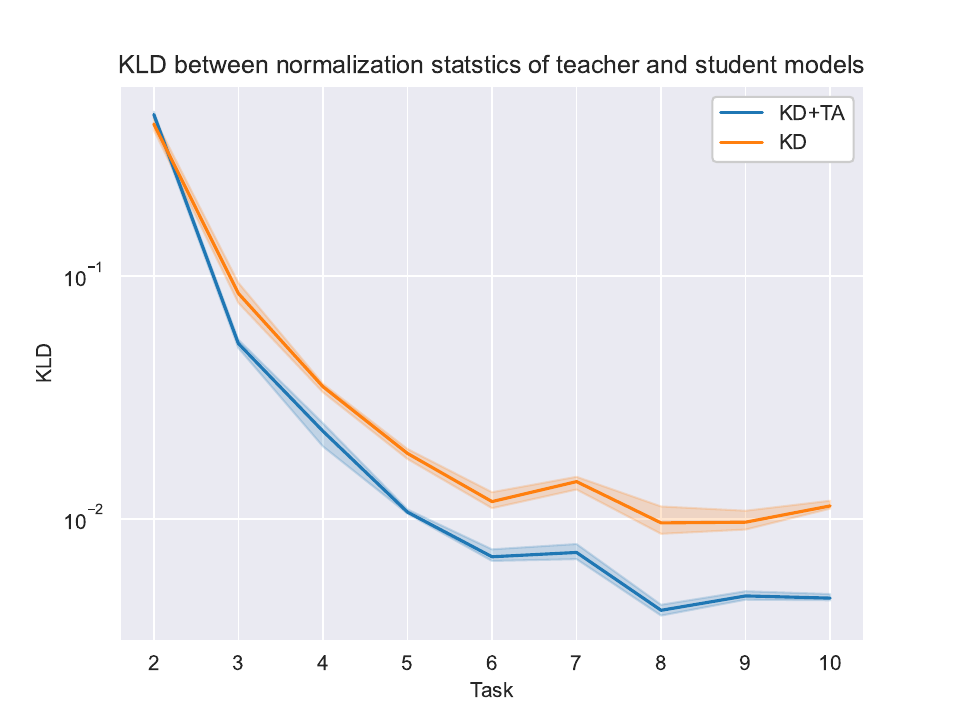}
    \end{subfigure}
    \begin{subfigure}[b]{0.49\textwidth}
        \centering
        \includegraphics[width=\textwidth]{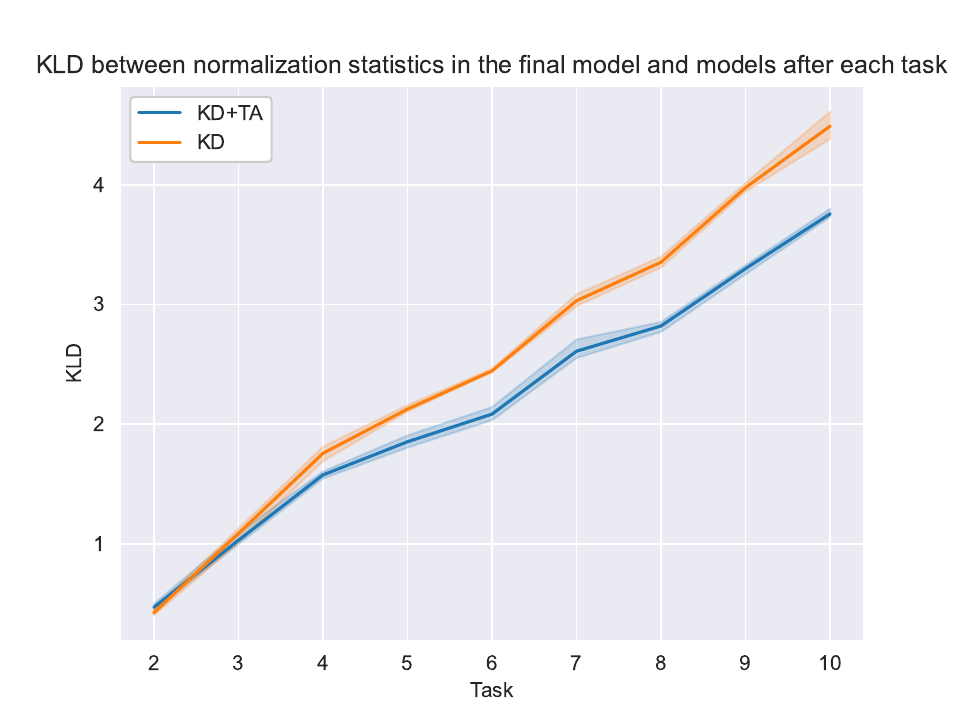}
    \end{subfigure}
   \caption{TA impact on batch normalization statistics. Left: KLD between the normalization statistics of the teacher and the student at the end of the training of all the tasks. Right: KLD between the normalization statistics of the model at the end of each task and the model trained on the first task.}
   \label{fig:kld}
\end{figure*}

We investigate the batch normalization statistics in the teacher and student model throughout CIL on CIFAR100 split into 10 equally sized tasks. To measure the divergence between the normalization statistics in the models, we compute the average Kullback-Leibler divergence (KLD) between the distributions of normalization statistics in every batch normalization layer. 
We show the results of this analysis in the \Cref{fig:kld}. Specifically, we measure KLD between the final teacher and student model at the end of each task, as well as the difference between the final student model trained for each task and the model initially learned after the first task. Applying TA leads to reduced KLD between the statistics in both teacher and student, and also student and initial model. This proves that our method leads to more stable representations throughout the training.

\begin{figure*}[t]
    \centering
    \begin{subfigure}[b]{0.49\textwidth}
        \centering
        \includegraphics[width=\textwidth]{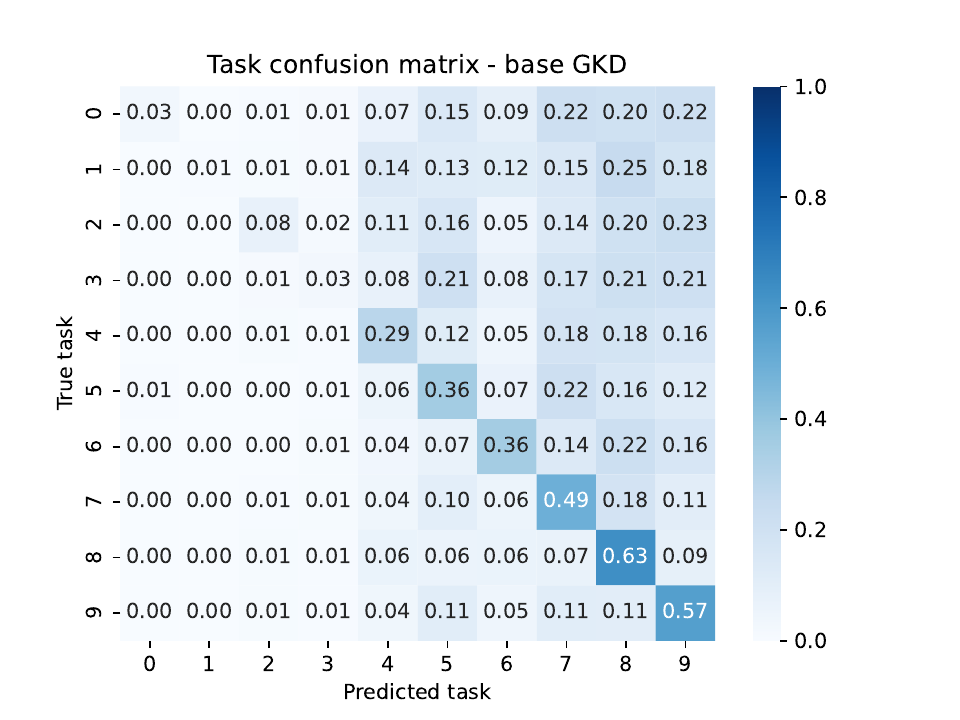}
          \end{subfigure}
    \begin{subfigure}[b]{0.49\textwidth}
        \centering
        \includegraphics[width=\textwidth]{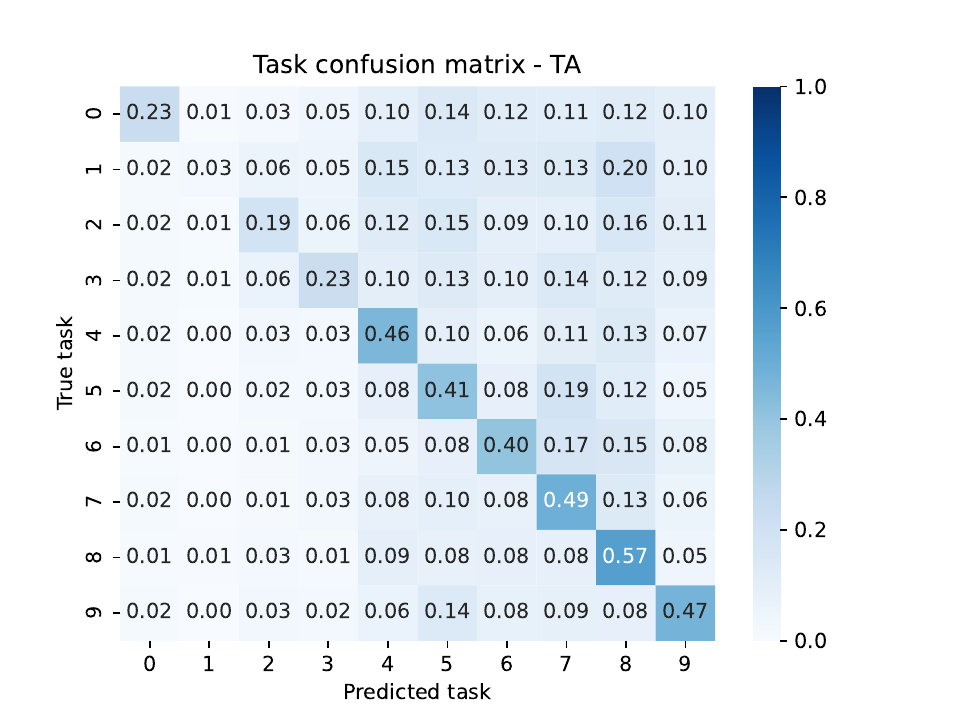}
    \end{subfigure}

   \caption{Task confusion matrix after learning all ten tasks on CIFAR100/10 for (left) base GKD and (right) GKD+TA. We see that TA leads to a model that is better at distinguishing between the first tasks and exhibits lower recency bias.}
   \label{fig:app:tcm}
\end{figure*}

\subsection{Hyperparameters for alternative adaptation methods ablation}
The best results reported in \Cref{tab:ta_ablations} were obtained with teacher learning rate of 1e-7. For variants with teacher pertaining, the best results were obtained with training for 5 epochs. We keep $\lambda$ fixed at 10 for all methods reported in the Table. While all variants that allow adaptation of batch norm statistics improve upon the baseline, the values of best hyperparameters are generally small, suggesting that the main source of improvement in all the methods comes from changes in normalization. This is further supported by the fact that all improvements vanish when normalization statistics are fixed.

\subsection{Task recency bias with TA}
We also conduct additional analysis of our method of Teacher Adaptation (TA) to understand the mechanism with which it improves upon the standard knowledge distillation. At \Cref{fig:app:tcm}, we analyze task confusion matrices of standard knowledge distillation (LwF) and its extension with TA.  We find that applying TA results in a model that is better at distinguishing between the tasks, and generally exhibits lower recency bias. 

We hypothesize that the lower KD loss that we observe when using TA results in smaller updates to the model, so the difference between the magnitudes of logits learned for different tasks is smaller. Therefore, Teacher Adaptation helps to alleviate the recency bias in class-incremental learning.

\section{Additional experiments for different CL benchmarks}
\label{sec:app:experiments}
We provide the results of additional experiments conducted with our method and warmup. To shorten the notation, we denote the total number of tasks, including the first pretraining task (if present), with \textbf{T}, and the number of classes in the first task by \textbf{S}. For example, for CIFAR100 \textbf{T10S10} is a setting composed of 10 tasks with 10 classes each, while \textbf{T11S50} is a setting where the first task contains 50 classes and the next 10 tasks contain 5 classes each.

\subsection{Standard benchmarks}
In addition to \Cref{sec:exp:benchmarks}, we conduct more experiments on CIFAR100, TinyImageNet200 and ImageNet100, adding two settings with a smaller number of tasks. We report final accuracy and forgetting in addition to incremental accuracy and forgetting. Additionally, we include the results for the warmup phase (WU) applied in isolation, without the Teacher Adaptation.

We report the results for CIFAR100, TinyImageNet200 and ImageNet100 in \Cref{tab:app:cifar100}, \Cref{tab:app:tin200}, and \Cref{tab:app:imagenet100} respectively.

\begin{table*}[!th]
\centering
\resizebox{0.95\textwidth}{!}{%
\begin{tabular}{@{}lrrrrrrrr@{}}
\toprule
 &
  \multicolumn{4}{c}{T5S20} &
  \multicolumn{4}{c}{T6S50} \\ \cmidrule{2-9}
\textit{} &
  \multicolumn{1}{c}{$Acc_{Final} \uparrow$} &
  \multicolumn{1}{c}{$Acc_{Inc} \uparrow$} &
  \multicolumn{1}{c}{$Forg_{Final} \downarrow$} &
  \multicolumn{1}{c}{$Forg_{Inc} \downarrow$} &
  \multicolumn{1}{c}{$Acc_{Final} \uparrow$} &
  \multicolumn{1}{c}{$Acc_{Inc} \uparrow$} &
  \multicolumn{1}{c}{$Forg_{Final} \downarrow$} &
  \multicolumn{1}{c}{$Forg_{Inc} \downarrow$} \\ \midrule
GKD &
  37.63$\pm$0.52 &
  48.80$\pm$0.36 &
  23.13$\pm$2.28 &
  19.30$\pm$2.37 &
  40.74$\pm$0.72 &
  51.93$\pm$1.24 &
  25.87$\pm$0.75 &
  18.90$\pm$0.29 \\
\textit{+TA} &
  40.84$\pm$0.23 &
  50.08$\pm$0.26 &
  \textbf{16.76$\pm$1.68} &
  \textbf{16.66$\pm$1.25} &
  43.18$\pm$1.66 &
  52.22$\pm$1.28 &
  \textbf{18.73$\pm$1.61} &
  \textbf{14.09$\pm$0.66} \\
\textit{+WU} &
  38.82$\pm$0.06 &
  49.69$\pm$0.26 &
  22.08$\pm$0.61 &
  19.34$\pm$1.24 &
  43.19$\pm$0.95 &
  \textbf{53.42$\pm$0.50} &
  24.54$\pm$2.17 &
  18.94$\pm$2.69 \\
\textit{+TA+WU} &
  \textbf{41.50$\pm$0.60} &
  \textbf{50.70$\pm$0.27} &
  16.85$\pm$0.74 &
  17.16$\pm$1.24 &
  \textbf{44.66$\pm$0.54} &
  53.40$\pm$0.73 &
  \textbf{18.73$\pm$1.73} &
  14.53$\pm$1.56 \\ \midrule
MKD &
  34.41$\pm$0.16 &
  47.42$\pm$0.32 &
  23.74$\pm$1.63 &
  23.51$\pm$1.75 &
  39.33$\pm$0.72 &
  50.28$\pm$1.12 &
  23.54$\pm$0.62 &
  18.19$\pm$0.31 \\
\textit{+TA} &
  37.03$\pm$0.82 &
  48.63$\pm$0.17 &
  \textbf{14.64$\pm$0.22} &
  \textbf{17.63$\pm$0.73} &
  39.45$\pm$1.29 &
  49.42$\pm$1.38 &
  \textbf{13.97$\pm$0.89} &
  \textbf{11.02$\pm$0.28} \\
\textit{+WU} &
  35.37$\pm$0.19 &
  47.96$\pm$0.15 &
  24.63$\pm$0.81 &
  24.35$\pm$0.44 &
  40.89$\pm$1.52 &
  \textbf{51.46$\pm$0.68} &
  24.45$\pm$2.59 &
  19.53$\pm$2.86 \\
\textit{+TA+WU} &
  \textbf{38.26$\pm$0.24} &
  \textbf{49.58$\pm$0.06} &
  15.18$\pm$0.41 &
  17.96$\pm$0.65 &
  \textbf{41.42$\pm$0.78} &
  51.04$\pm$1.17 &
  15.00$\pm$1.07 &
  12.03$\pm$1.32 \\ \midrule
TKD &
  38.33$\pm$0.70 &
  49.56$\pm$0.48 &
  24.78$\pm$2.58 &
  25.04$\pm$2.55 &
  41.19$\pm$0.42 &
  52.07$\pm$1.35 &
  17.31$\pm$1.20 &
  15.02$\pm$0.39 \\
\textit{+TA} &
  41.12$\pm$0.35 &
  50.87$\pm$0.15 &
  18.12$\pm$1.55 &
  \textbf{21.09$\pm$1.39} &
  41.36$\pm$0.89 &
  51.88$\pm$0.80 &
  \textbf{12.84$\pm$1.25} &
  \textbf{11.42$\pm$0.64} \\
\textit{+WU} &
  40.03$\pm$0.66 &
  50.50$\pm$0.17 &
  22.79$\pm$0.73 &
  24.19$\pm$1.61 &
  \textbf{43.16$\pm$0.26} &
  \textbf{53.51$\pm$0.39} &
  18.40$\pm$1.73 &
  16.03$\pm$2.77 \\
\textit{+TA+WU} &
  \textbf{42.07$\pm$0.67} &
  \textbf{51.47$\pm$0.33} &
  \textbf{17.77$\pm$1.37} &
  21.21$\pm$1.34 &
  43.08$\pm$0.35 &
  53.12$\pm$0.60 &
  13.64$\pm$0.89 &
  12.13$\pm$1.37 \\
  
  \midrule
  ANCL &
  37.13$\pm$0.33 & 48.60$\pm$0.27 & 32.41$\pm$0.26 & 31.71$\pm$0.81 & 43.41$\pm$0.62 & 53.40$\pm$0.49 & 23.54$\pm$1.29 & 18.08$\pm$0.40 \\
  \textit{+TA} &
  41.04$\pm$0.22 & 50.36$\pm$0.43 & 24.07$\pm$0.81 & 26.18$\pm$0.51 & 44.64$\pm$1.65 & 53.33$\pm$0.66 & 18.43$\pm$3.12 & 14.44$\pm$1.29 \\
  \textit{+WU} &
  38.94$\pm$0.57 & 49.26$\pm$0.61 & 29.84$\pm$0.88 & 30.98$\pm$1.98 & 44.74$\pm$0.30 & \textbf{54.25$\pm$0.29} & 22.42$\pm$1.75 & 17.49$\pm$1.56 \\
  \textit{+TA+WU} &
  \textbf{42.08$\pm$0.53} & \textbf{50.59$\pm$0.21} & \textbf{22.55$\pm$1.15} & \textbf{26.09$\pm$1.11} & \textbf{45.64$\pm$0.87} & 54.05$\pm$0.42 & \textbf{18.12$\pm$1.58} & \textbf{14.28$\pm$1.06} \\
  \midrule
 &
  \multicolumn{4}{c}{T10S10} &
  \multicolumn{4}{c}{T11S50} \\ 
  \midrule
GKD &
  28.27$\pm$0.44 &
  42.52$\pm$0.76 &
  29.59$\pm$0.92 &
  22.26$\pm$0.31 &
  30.79$\pm$1.62 &
  41.69$\pm$1.18 &
  26.84$\pm$2.12 &
  18.09$\pm$0.88 \\
\textit{+TA} &
  31.92$\pm$0.86 &
  44.09$\pm$0.97 &
  22.65$\pm$1.32 &
  \textbf{19.41$\pm$0.60} &
  33.20$\pm$0.76 &
  44.05$\pm$1.12 &
  \textbf{18.90$\pm$0.19} &
  \textbf{12.97$\pm$0.43} \\
\textit{+WU} &
  31.11$\pm$0.58 &
  43.95$\pm$0.65 &
  25.26$\pm$0.93 &
  21.95$\pm$0.74 &
  33.17$\pm$0.86 &
  44.09$\pm$1.26 &
  23.85$\pm$1.29 &
  17.25$\pm$0.74 \\
\textit{+TA+WU} &
  \textbf{33.76$\pm$0.78} &
  \textbf{45.25$\pm$1.02} &
  \textbf{20.87$\pm$0.13} &
  19.87$\pm$0.34 &
  \textbf{34.72$\pm$0.77} &
  \textbf{46.27$\pm$1.09} &
  19.13$\pm$1.77 &
  13.98$\pm$0.98 \\ \midrule
MKD &
  25.41$\pm$1.07 &
  39.36$\pm$0.70 &
  44.04$\pm$0.98 &
  42.74$\pm$0.52 &
  31.31$\pm$1.22 &
  41.04$\pm$0.93 &
  22.56$\pm$0.94 &
  15.37$\pm$0.33 \\
\textit{+TA} &
  29.97$\pm$1.44 &
  43.55$\pm$0.96 &
  32.37$\pm$2.17 &
  32.61$\pm$0.80 &
  30.90$\pm$0.69 &
  41.67$\pm$1.35 &
  \textbf{14.76$\pm$0.25} &
  \textbf{10.51$\pm$0.36} \\
\textit{+WU} &
  27.32$\pm$1.28 &
  40.78$\pm$0.95 &
  37.89$\pm$0.22 &
  39.03$\pm$0.53 &
  31.63$\pm$1.16 &
  42.45$\pm$0.78 &
  22.57$\pm$1.57 &
  16.98$\pm$1.23 \\
\textit{+TA+WU} &
  \textbf{31.94$\pm$0.23} &
  \textbf{44.85$\pm$0.80} &
  \textbf{26.83$\pm$1.01} &
  \textbf{30.08$\pm$0.57} &
  \textbf{32.79$\pm$0.55} &
  \textbf{44.19$\pm$1.17} &
  16.44$\pm$0.63 &
  12.56$\pm$0.53 \\
  \midrule
TKD &
  30.05$\pm$0.81 &
  43.74$\pm$0.84 &
  24.53$\pm$0.23 &
  23.65$\pm$0.79 &
  28.38$\pm$1.46 &
  40.44$\pm$1.40 &
  15.68$\pm$0.84 &
  12.20$\pm$0.46 \\
\textit{+TA} &
  31.80$\pm$0.67 &
  45.29$\pm$1.02 &
  18.59$\pm$0.90 &
  \textbf{19.42$\pm$0.85} &
  28.50$\pm$0.39 &
  41.68$\pm$1.03 &
  \textbf{11.58$\pm$0.38} &
  \textbf{9.29$\pm$0.75} \\
\textit{+WU} &
  31.23$\pm$0.94 &
  44.59$\pm$0.72 &
  23.21$\pm$0.35 &
  23.73$\pm$0.32 &
  \textbf{30.88$\pm$1.16} &
  42.45$\pm$1.31 &
  17.94$\pm$0.39 &
  14.30$\pm$0.99 \\
\textit{+TA+WU} &
  \textbf{33.14$\pm$1.03} &
  \textbf{46.21$\pm$0.86} &
  \textbf{17.55$\pm$0.66} &
  20.45$\pm$0.57 &
  30.60$\pm$0.37 &
  \textbf{44.22$\pm$1.08} &
  14.65$\pm$0.49 &
  11.79$\pm$0.57 \\

  \midrule
  ANCL &
  29.77$\pm$1.10 & 43.15$\pm$0.49 & 36.37$\pm$1.48 & 32.78$\pm$1.52 & 29.77$\pm$1.10  & 43.15$\pm$0.49 & 36.37$\pm$1.48 & 32.78$\pm$1.52 \\
  \textit{+TA} &
  33.47$\pm$0.17 & 45.67$\pm$0.14 & \textbf{28.04$\pm$1.61} & \textbf{26.71$\pm$1.71} & 33.47$\pm$0.17 & 45.67$\pm$0.14 & \textbf{28.04$\pm$1.61} & \textbf{26.71$\pm$1.71} \\
  \textit{+WU} &
  31.41$\pm$0.54 & 44.28$\pm$0.08 & 34.70$\pm$0.25  & 32.06$\pm$1.07 & 31.41$\pm$0.54 & 44.28$\pm$0.08 & 34.70$\pm$0.25  & 32.06$\pm$1.07 \\
  \textit{+TA+WU} &
  \textbf{34.17$\pm$0.30}  & \textbf{46.73$\pm$0.20}  & 28.67$\pm$0.63 & 26.86$\pm$1.17 & \textbf{34.17$\pm$0.30}  & \textbf{46.73$\pm$0.20}  & 28.67$\pm$0.63 & 26.86$\pm$1.17 \\
  \midrule
  
 &
  \multicolumn{4}{c}{T20S5} &
  \multicolumn{4}{c}{T26S50} \\ 
  \midrule
GKD &
  15.59$\pm$0.32 &
  31.89$\pm$0.45 &
  43.28$\pm$0.56 &
  34.68$\pm$1.87 &
  10.10$\pm$0.71 &
  17.64$\pm$0.93 &
  15.29$\pm$0.27 &
  9.67$\pm$0.26 \\
\textit{+TA} &
  19.55$\pm$0.24 &
  35.99$\pm$0.79 &
  30.38$\pm$2.08 &
  \textbf{23.32$\pm$1.79} &
  11.99$\pm$0.66 &
  19.37$\pm$1.73 &
  \textbf{9.05$\pm$0.63} &
  \textbf{8.31$\pm$0.68} \\
\textit{+WU} &
  19.22$\pm$0.80 &
  34.67$\pm$0.55 &
  37.05$\pm$1.03 &
  31.14$\pm$1.41 &
  10.41$\pm$1.01 &
  21.87$\pm$0.16 &
  17.92$\pm$0.48 &
  11.02$\pm$0.65 \\
\textit{+TA+WU} &
  \textbf{21.52$\pm$0.62} &
  \textbf{37.11$\pm$0.64} &
  \textbf{30.34$\pm$0.78} &
  24.87$\pm$1.04 &
  \textbf{18.11$\pm$0.52} &
  \textbf{26.15$\pm$0.94} &
  10.11$\pm$0.97 &
  8.73$\pm$0.85 \\ \midrule
MKD &
  17.61$\pm$0.51 &
  32.89$\pm$0.42 &
  33.36$\pm$0.80 &
  32.01$\pm$1.36 &
  11.55$\pm$0.86 &
  19.14$\pm$1.36 &
  14.04$\pm$0.71 &
  8.76$\pm$0.35 \\
\textit{+TA} &
  19.07$\pm$0.65 &
  35.36$\pm$0.85 &
  \textbf{21.27$\pm$1.08} &
  \textbf{20.96$\pm$1.31} &
  13.91$\pm$0.80 &
  20.99$\pm$1.53 &
  \textbf{9.52$\pm$0.26} &
  \textbf{8.20$\pm$0.98} \\
\textit{+WU} &
  18.96$\pm$0.82 &
  34.30$\pm$0.94 &
  30.45$\pm$0.70 &
  29.75$\pm$1.56 &
  11.48$\pm$0.36 &
  22.42$\pm$0.76 &
  17.42$\pm$1.70 &
  11.51$\pm$1.19 \\
\textit{+TA+WU} &
  \textbf{20.52$\pm$0.85} &
  \textbf{36.79$\pm$0.70} &
  21.88$\pm$0.81 &
  21.84$\pm$0.33 &
  \textbf{18.15$\pm$0.70} &
  \textbf{26.10$\pm$0.75} &
  11.64$\pm$1.04 &
  9.17$\pm$0.20 \\ \midrule
TKD &
  19.39$\pm$0.41 &
  34.58$\pm$0.34 &
  22.06$\pm$0.46 &
  21.13$\pm$1.17 &
  7.88$\pm$0.08 &
  14.64$\pm$0.33 &
  7.96$\pm$0.47 &
  \textbf{6.02$\pm$0.54} \\
\textit{+TA} &
  18.30$\pm$0.50 &
  34.62$\pm$0.92 &
  \textbf{15.22$\pm$1.25} &
  \textbf{14.72$\pm$1.28} &
  9.05$\pm$0.64 &
  16.66$\pm$1.66 &
  \textbf{7.17$\pm$0.53} &
  6.88$\pm$0.36 \\
\textit{+WU} &
  \textbf{20.77$\pm$0.59} &
  35.72$\pm$0.09 &
  21.14$\pm$0.88 &
  20.93$\pm$1.62 &
  9.55$\pm$0.53 &
  18.01$\pm$0.59 &
  12.63$\pm$1.04 &
  9.90$\pm$0.99 \\
\textit{+TA+WU} &
  20.24$\pm$0.97 &
  \textbf{36.26$\pm$0.71} &
  17.98$\pm$1.00 &
  17.01$\pm$0.89 &
  \textbf{13.34$\pm$0.73} &
  \textbf{22.00$\pm$0.97} &
  10.80$\pm$1.03 &
  9.36$\pm$0.68 \\ 

  \midrule
  ANCL &
  19.21$\pm$0.75 & 34.32$\pm$0.41 & 39.97$\pm$1.81 & 36.74$\pm$1.38 & 11.04$\pm$0.40  & 21.84$\pm$1.33 & 18.31$\pm$1.64 & 11.79$\pm$0.51 \\
  \textit{+TA} &
  20.63$\pm$1.04 & 37.52$\pm$0.78 & \textbf{31.07$\pm$1.00}  & \textbf{26.77$\pm$0.49} & 18.96$\pm$0.91 & 26.47$\pm$0.41 & \textbf{12.69$\pm$1.49} & \textbf{10.33$\pm$0.57} \\
  \textit{+WU} &
  21.53$\pm$0.38 & 36.06$\pm$0.73 & 37.34$\pm$0.53 & 35.34$\pm$1.52 & 11.81$\pm$0.24 & 23.81$\pm$0.88 & 20.05$\pm$0.95 & 13.18$\pm$0.34 \\
  \textit{+TA+WU} &
  \textbf{22.54$\pm$0.33} & \textbf{38.48$\pm$0.94} & 31.87$\pm$0.92 & 27.99$\pm$1.20  & \textbf{20.79$\pm$0.28} & \textbf{29.67$\pm$1.13} & 13.84$\pm$0.45 & 10.98$\pm$0.56 \\

  \bottomrule
\end{tabular}%
}
\caption{Additional results for CIFAR100}
\label{tab:app:cifar100}
\end{table*}

\begin{table*}[!th]
\centering
\resizebox{0.95\textwidth}{!}{%
\begin{tabular}{lrrrrrrrr}
\toprule
 &
  \multicolumn{4}{c}{T5S40} &
  \multicolumn{4}{c}{T6S100} \\ \cmidrule{2-9}
 &
    \multicolumn{1}{c}{$Acc_{Final} \uparrow$} &
  \multicolumn{1}{c}{$Acc_{Inc} \uparrow$} &
  \multicolumn{1}{c}{$Forg_{Final} \downarrow$} &
  \multicolumn{1}{c}{$Forg_{Inc} \downarrow$} &
  \multicolumn{1}{c}{$Acc_{Final} \uparrow$} &
  \multicolumn{1}{c}{$Acc_{Inc} \uparrow$} &
  \multicolumn{1}{c}{$Forg_{Final} \downarrow$} &
  \multicolumn{1}{c}{$Forg_{Inc} \downarrow$} \\ 
  \midrule
GKD &
  27.51$\pm$0.81 &
  37.53$\pm$1.28 &
  18.22$\pm$0.55 &
  14.58$\pm$0.56 &
  32.92$\pm$1.40 &
  38.74$\pm$2.77 &
  13.77$\pm$2.12 &
  9.60$\pm$2.04 \\
\textit{+TA} &
  28.36$\pm$1.13 &
  37.60$\pm$1.49 &
  15.55$\pm$1.79 &
  12.63$\pm$0.39 &
  33.79$\pm$0.89 &
  38.56$\pm$1.73 &
  10.94$\pm$0.71 &
  \textbf{8.28$\pm$0.73} \\
\textit{+WU} &
  28.31$\pm$0.66 &
  37.74$\pm$1.00 &
  17.32$\pm$0.54 &
  15.04$\pm$0.93 &
  33.80$\pm$0.79 &
  39.33$\pm$1.84 &
  13.04$\pm$2.02 &
  9.44$\pm$1.29 \\
\textit{+TA+WU} &
  \textbf{29.36$\pm$0.81} &
  \textbf{38.21$\pm$1.10} &
  \textbf{14.05$\pm$1.38} &
  \textbf{12.50$\pm$0.60} &
  \textbf{34.68$\pm$1.06} &
  \textbf{39.45$\pm$1.73} &
  \textbf{10.40$\pm$0.73} &
  8.36$\pm$0.78 \\ \midrule
MKD &
  24.09$\pm$0.94 &
  34.93$\pm$1.39 &
  11.53$\pm$0.17 &
  10.22$\pm$1.34 &
  29.65$\pm$0.77 &
  35.48$\pm$2.06 &
  9.27$\pm$1.86 &
  6.66$\pm$1.30 \\
\textit{+TA} &
  24.17$\pm$0.77 &
  34.67$\pm$1.58 &
  6.57$\pm$0.26 &
  \textbf{6.51$\pm$0.67} &
  28.28$\pm$0.79 &
  34.07$\pm$1.63 &
  \textbf{4.31$\pm$0.47} &
  \textbf{4.14$\pm$0.60} \\
\textit{+WU} &
  \textbf{24.55$\pm$0.73} &
  \textbf{35.28$\pm$1.08} &
  11.10$\pm$0.07 &
  10.79$\pm$1.45 &
  \textbf{30.40$\pm$1.05} &
  \textbf{36.43$\pm$2.08} &
  9.79$\pm$2.19 &
  7.10$\pm$1.53 \\
\textit{+TA+WU} &
  24.33$\pm$0.46 &
  35.05$\pm$1.37 &
  \textbf{6.28$\pm$0.70} &
  6.92$\pm$1.04 &
  28.53$\pm$1.58 &
  34.81$\pm$1.95 &
  4.49$\pm$0.34 &
  4.42$\pm$0.63 \\ \midrule
TKD &
  28.18$\pm$0.92 &
  37.43$\pm$0.86 &
  19.76$\pm$1.11 &
  20.37$\pm$0.78 &
  34.17$\pm$0.72 &
  40.11$\pm$1.59 &
  12.33$\pm$1.05 &
  11.59$\pm$1.22 \\
\textit{+TA} &
  28.95$\pm$0.93 &
  38.07$\pm$1.03 &
  16.36$\pm$1.91 &
  17.23$\pm$0.56 &
  34.42$\pm$0.61 &
  40.20$\pm$1.31 &
  10.13$\pm$1.03 &
  10.15$\pm$1.03 \\
\textit{+WU} &
  29.07$\pm$0.54 &
  37.73$\pm$0.71 &
  17.35$\pm$1.16 &
  19.68$\pm$0.80 &
  34.22$\pm$0.79 &
  40.44$\pm$1.63 &
  12.09$\pm$1.68 &
  11.48$\pm$1.58 \\
\textit{+TA+WU} &
  \textbf{29.57$\pm$0.46} &
  \textbf{38.29$\pm$0.74} &
  \textbf{14.69$\pm$1.34} &
  \textbf{16.93$\pm$0.86} &
  \textbf{34.63$\pm$1.07} &
  \textbf{40.64$\pm$1.38} &
  \textbf{9.55$\pm$0.60} &
  \textbf{10.01$\pm$1.07} \\ \midrule
LwF(A) &
  28.33$\pm$0.36 &
  37.63$\pm$0.83 &
  21.80$\pm$1.09 &
  20.57$\pm$0.85 &
  34.48$\pm$0.78 &
  40.85$\pm$1.49 &
  19.51$\pm$0.94 &
  15.69$\pm$1.43 \\
\textit{+TA} &
  29.37$\pm$0.69 &
  38.31$\pm$1.06 &
  19.34$\pm$1.20 &
  \textbf{18.17$\pm$0.74} &
  36.01$\pm$1.02 &
  41.22$\pm$1.46 &
  \textbf{15.83$\pm$0.35} &
  \textbf{13.75$\pm$1.12} \\
\textit{+WU} &
  28.59$\pm$0.63 &
  37.69$\pm$0.73 &
  21.70$\pm$0.96 &
  20.87$\pm$1.02 &
  34.64$\pm$0.38 &
  41.11$\pm$1.38 &
  20.23$\pm$1.55 &
  15.70$\pm$1.97 \\
\textit{+TA+WU} &
  \textbf{29.98$\pm$0.86} &
  \textbf{38.45$\pm$0.84} &
  \textbf{18.66$\pm$1.61} &
  18.35$\pm$0.98 &
  \textbf{36.20$\pm$0.73} &
  \textbf{41.54$\pm$1.21} &
  16.64$\pm$1.02 &
  13.85$\pm$1.43 \\ \midrule
 &
  \multicolumn{4}{c}{T10S20} &
  \multicolumn{4}{c}{T11S100} \\ 
  \midrule
GKD &
  20.76$\pm$0.96 &
  32.12$\pm$0.57 &
  26.87$\pm$2.16 &
  21.21$\pm$1.31 &
  27.17$\pm$0.73 &
  34.32$\pm$1.96 &
  17.53$\pm$3.99 &
  11.41$\pm$2.80 \\
\textit{+TA} &
  22.94$\pm$0.70 &
  33.12$\pm$0.53 &
  21.39$\pm$2.14 &
  \textbf{17.50$\pm$1.76} &
  30.65$\pm$0.55 &
  37.01$\pm$1.84 &
  \textbf{12.85$\pm$1.60} &
  \textbf{10.05$\pm$0.87} \\
\textit{+WU} &
  21.94$\pm$0.71 &
  33.00$\pm$0.59 &
  24.84$\pm$1.49 &
  20.37$\pm$0.65 &
  28.77$\pm$0.81 &
  36.54$\pm$1.45 &
  18.11$\pm$1.90 &
  11.66$\pm$1.83 \\
\textit{+TA+WU} &
  \textbf{23.72$\pm$1.34} &
  \textbf{33.90$\pm$0.78} &
  \textbf{21.29$\pm$2.19} &
  17.70$\pm$1.48 &
  \textbf{31.34$\pm$0.49} &
  \textbf{37.50$\pm$1.84} &
  13.25$\pm$2.11 &
  10.14$\pm$1.10 \\ \midrule
MKD &
  19.37$\pm$0.85 &
  31.04$\pm$1.00 &
  17.04$\pm$0.66 &
  16.90$\pm$0.75 &
  26.11$\pm$0.72 &
  33.75$\pm$2.11 &
  16.22$\pm$2.18 &
  10.30$\pm$1.63 \\
\textit{+TA} &
  20.74$\pm$0.88 &
  31.78$\pm$0.83 &
  \textbf{9.89$\pm$1.58} &
  \textbf{11.42$\pm$1.23} &
  25.59$\pm$1.25 &
  32.23$\pm$2.28 &
  \textbf{6.18$\pm$0.99} &
  \textbf{5.91$\pm$0.66} \\
\textit{+WU} &
  19.93$\pm$0.67 &
  31.32$\pm$0.95 &
  16.81$\pm$1.12 &
  16.97$\pm$0.27 &
  26.36$\pm$0.80 &
  \textbf{33.91$\pm$1.39} &
  17.12$\pm$2.33 &
  11.17$\pm$1.71 \\
\textit{+TA+WU} &
  \textbf{21.42$\pm$0.75} &
  \textbf{32.22$\pm$0.95} &
  10.13$\pm$1.25 &
  11.94$\pm$1.07 &
  \textbf{26.79$\pm$1.56} &
  32.99$\pm$1.98 &
  6.78$\pm$0.92 &
  6.42$\pm$0.80 \\ \midrule
TKD &
  22.88$\pm$0.29 &
  33.15$\pm$0.45 &
  19.17$\pm$0.02 &
  21.05$\pm$0.39 &
  28.32$\pm$1.18 &
  37.31$\pm$1.36 &
  14.34$\pm$0.78 &
  11.83$\pm$1.31 \\
\textit{+TA} &
  23.92$\pm$0.49 &
  33.94$\pm$0.70 &
  \textbf{14.91$\pm$0.98} &
  17.32$\pm$0.97 &
  30.18$\pm$0.92 &
  37.97$\pm$1.47 &
  \textbf{7.93$\pm$0.77} &
  \textbf{9.02$\pm$0.65} \\
\textit{+WU} &
  23.32$\pm$0.61 &
  33.71$\pm$0.97 &
  18.77$\pm$1.33 &
  20.33$\pm$0.41 &
  27.90$\pm$2.05 &
  37.22$\pm$1.60 &
  14.43$\pm$1.16 &
  11.48$\pm$0.86 \\
\textit{+TA+WU} &
  \textbf{24.69$\pm$0.89} &
  \textbf{34.58$\pm$0.96} &
  15.09$\pm$0.92 &
  \textbf{17.27$\pm$0.54} &
  \textbf{30.80$\pm$0.75} &
  \textbf{38.41$\pm$1.52} &
  8.55$\pm$1.06 &
  9.16$\pm$0.62 \\ \midrule
ANCL &
  22.58$\pm$0.83 &
  32.84$\pm$0.78 &
  28.76$\pm$2.52 &
  27.24$\pm$1.47 &
  27.78$\pm$0.96 &
  37.74$\pm$0.60 &
  25.24$\pm$2.77 &
  17.28$\pm$2.59 \\
\textit{+TA} &
  23.98$\pm$0.68 &
  33.85$\pm$0.55 &
  \textbf{25.23$\pm$1.31} &
  24.02$\pm$1.39 &
  31.34$\pm$0.34 &
  39.50$\pm$1.18 &
  20.36$\pm$1.54 &
  \textbf{14.88$\pm$1.41} \\
\textit{+WU} &
  23.20$\pm$0.57 &
  33.33$\pm$0.81 &
  28.66$\pm$1.18 &
  26.86$\pm$0.20 &
  29.05$\pm$0.75 &
  38.86$\pm$1.29 &
  25.35$\pm$0.64 &
  17.11$\pm$0.89 \\
\textit{+TA+WU} &
  \textbf{24.78$\pm$0.86} &
  \textbf{34.59$\pm$0.75} &
  25.24$\pm$1.47 &
  \textbf{23.64$\pm$0.76} &
  \textbf{32.36$\pm$0.60} &
  \textbf{40.10$\pm$1.39} &
  \textbf{20.31$\pm$1.73} &
  15.12$\pm$1.01 \\ \midrule
 &
  \multicolumn{4}{c}{T20S10} &
  \multicolumn{4}{c}{T26S100} \\  
  \midrule
GKD &
  13.83$\pm$0.25 &
  25.75$\pm$0.65 &
  33.70$\pm$0.54 &
  27.55$\pm$1.81 &
  14.33$\pm$1.42 &
  23.30$\pm$1.59 &
  23.27$\pm$2.88 &
  12.63$\pm$2.21 \\
\textit{+TA} &
  16.15$\pm$0.83 &
  27.16$\pm$0.87 &
  27.52$\pm$0.51 &
  22.62$\pm$2.03 &
  19.70$\pm$0.79 &
  26.88$\pm$1.62 &
  \textbf{14.57$\pm$1.00} &
  \textbf{11.08$\pm$0.47} \\ 
\textit{+WU} &
  15.20$\pm$0.46 &
  26.98$\pm$1.05 &
  30.78$\pm$1.84 &
  25.57$\pm$1.97 &
  16.10$\pm$1.03 &
  25.59$\pm$1.76 &
  21.45$\pm$1.77 &
  12.66$\pm$1.65 \\ 
\textit{+TA+WU} &
  \textbf{17.65$\pm$0.16} &
  \textbf{27.85$\pm$0.90} &
  \textbf{25.60$\pm$0.40} &
  \textbf{21.65$\pm$1.43} &
  \textbf{21.86$\pm$1.34} &
  \textbf{28.82$\pm$1.95} &
  15.56$\pm$0.77 &
  11.44$\pm$0.61 \\ \midrule
MKD &
  13.81$\pm$0.53 &
  25.22$\pm$0.88 &
  24.61$\pm$1.91 &
  25.80$\pm$2.46 &
  15.99$\pm$1.72 &
  23.42$\pm$1.95 &
  18.15$\pm$3.02 &
  10.41$\pm$1.97 \\
\textit{+TA} &
  16.66$\pm$0.59 &
  27.05$\pm$1.11 &
  14.46$\pm$0.75 &
  17.55$\pm$2.36 &
  17.65$\pm$0.32 &
  23.29$\pm$1.61 &
  \textbf{8.57$\pm$1.70} &
  \textbf{7.61$\pm$1.18} \\
\textit{+WU} &
  13.97$\pm$0.62 &
  25.76$\pm$0.73 &
  24.33$\pm$2.37 &
  24.76$\pm$2.29 &
  16.74$\pm$1.44 &
  24.42$\pm$1.72 &
  19.95$\pm$2.21 &
  12.29$\pm$2.42 \\
\textit{+TA+WU} &
  \textbf{17.31$\pm$0.48} &
  \textbf{27.39$\pm$1.33} &
  \textbf{13.66$\pm$0.39} &
  \textbf{16.74$\pm$1.52} &
  \textbf{20.29$\pm$1.27} &
  \textbf{25.75$\pm$1.74} &
  9.48$\pm$1.04 &
  8.35$\pm$0.82 \\ \midrule
TKD &
  16.13$\pm$0.45 &
  27.29$\pm$0.76 &
  21.34$\pm$2.13 &
  23.20$\pm$2.46 &
  14.77$\pm$1.50 &
  23.94$\pm$2.14 &
  14.54$\pm$2.43 &
  10.20$\pm$1.09 \\
\textit{+TA} &
  18.34$\pm$0.25 &
  28.51$\pm$0.96 &
  14.67$\pm$0.26 &
  17.38$\pm$2.44 &
  18.20$\pm$0.34 &
  26.56$\pm$1.62 &
  \textbf{8.11$\pm$1.01} &
  \textbf{8.05$\pm$0.36} \\
\textit{+WU} &
  16.16$\pm$0.29 &
  28.02$\pm$0.63 &
  21.28$\pm$1.94 &
  22.07$\pm$1.48 &
  15.71$\pm$1.15 &
  24.74$\pm$2.16 &
  16.75$\pm$1.81 &
  12.14$\pm$2.04 \\
\textit{+TA+WU} &
  \textbf{18.99$\pm$0.50} &
  \textbf{28.71$\pm$1.06} &
  \textbf{14.58$\pm$0.94} &
  \textbf{17.37$\pm$1.78} &
  \textbf{20.07$\pm$0.87} &
  \textbf{28.10$\pm$1.97} &
  9.74$\pm$0.90 &
  9.56$\pm$0.65 \\ \midrule
ANCL &
  15.75$\pm$0.24 &
  26.98$\pm$0.60 &
  35.24$\pm$1.66 &
  32.45$\pm$1.18 &
  15.42$\pm$1.49 &
  27.95$\pm$1.98 &
  29.72$\pm$3.14 &
  20.91$\pm$0.92 \\
\textit{+TA} &
  17.91$\pm$0.33 &
  28.42$\pm$0.48 &
  30.13$\pm$1.18 &
  27.26$\pm$0.98 &
  21.04$\pm$1.79 &
  31.60$\pm$1.89 &
  23.59$\pm$1.71 &
  18.10$\pm$0.55 \\
\textit{+WU} &
  16.40$\pm$0.39 &
  27.54$\pm$0.44 &
  34.51$\pm$1.06 &
  31.76$\pm$0.91 &
  16.47$\pm$1.58 &
  28.80$\pm$1.59 &
  29.62$\pm$2.49 &
  20.74$\pm$1.10 \\
\textit{+TA+WU} &
  \textbf{19.02$\pm$0.34} &
  \textbf{29.18$\pm$0.71} &
  \textbf{28.78$\pm$0.40} &
  \textbf{26.50$\pm$1.34} &
  \textbf{22.32$\pm$1.25} &
  \textbf{32.60$\pm$1.45} &
  \textbf{23.21$\pm$1.01} &
  \textbf{17.94$\pm$0.39} \\ \bottomrule
\end{tabular}%
}
\caption{Additional results for TinyImageNet200.}
\label{tab:app:tin200}
\end{table*}

\begin{table*}[!th]
\centering
\resizebox{0.95\textwidth}{!}{%
\begin{tabular}{@{}lrrrrrrrr@{}}
\toprule
&
  \multicolumn{4}{c}{T5S20} &
  \multicolumn{4}{c}{T6S50} \\ \cmidrule{2-9}
 &
  \multicolumn{1}{c}{$Acc_{Final} \uparrow$} &
  \multicolumn{1}{c}{$Acc_{Inc} \uparrow$} &
  \multicolumn{1}{c}{$Forg_{Final} \downarrow$} &
  \multicolumn{1}{c}{$Forg_{Inc} \downarrow$} &
  \multicolumn{1}{c}{$Acc_{Final} \uparrow$} &
  \multicolumn{1}{c}{$Acc_{Inc} \uparrow$} &
  \multicolumn{1}{c}{$Forg_{Final} \downarrow$} &
  \multicolumn{1}{c}{$Forg_{Inc} \downarrow$} 
\\ \midrule
GKD &
  51.06$\pm$0.59 &
  61.95$\pm$0.50 &
  27.57$\pm$0.79 &
  23.16$\pm$0.88 &
  57.30$\pm$1.40 &
  66.36$\pm$0.75 &
  19.19$\pm$0.53 &
  12.92$\pm$0.59 \\
\textit{+TA} &
  \textbf{52.29$\pm$0.28} &
  \textbf{62.89$\pm$0.32} &
  \textbf{23.40$\pm$0.31} &
  \textbf{18.94$\pm$0.43} &
  57.86$\pm$0.68 &
  66.20$\pm$0.40 &
  \textbf{16.81$\pm$1.00} &
  \textbf{12.70$\pm$0.36} \\
\textit{+WU} &
  51.04$\pm$0.53 &
  60.24$\pm$0.23 &
  26.26$\pm$0.63 &
  27.22$\pm$0.29 &
  \textbf{59.52$\pm$0.27} &
  \textbf{67.63$\pm$0.39} &
  18.26$\pm$0.76 &
  14.45$\pm$0.47 \\
\textit{+TA+WU} &
  51.27$\pm$0.18 &
  61.01$\pm$0.12 &
  23.98$\pm$0.36 &
  23.51$\pm$0.29 &
  58.40$\pm$0.86 &
  66.86$\pm$0.31 &
  18.87$\pm$0.75 &
  15.29$\pm$0.69 \\ \midrule
MKD &
  50.17$\pm$0.25 &
  61.48$\pm$0.35 &
  23.21$\pm$0.78 &
  21.01$\pm$0.35 &
  54.70$\pm$1.02 &
  65.82$\pm$0.24 &
  27.21$\pm$0.70 &
  18.25$\pm$0.32 \\
\textit{+TA} &
  50.92$\pm$0.42 &
  62.18$\pm$0.45 &
  \textbf{16.00$\pm$0.61} &
  \textbf{14.71$\pm$0.35} &
  57.26$\pm$0.83 &
  66.03$\pm$0.32 &
  \textbf{19.00$\pm$1.04} &
  \textbf{14.34$\pm$0.11} \\
\textit{+WU} &
  50.45$\pm$0.60 &
  61.45$\pm$0.47 &
  23.20$\pm$0.50 &
  21.42$\pm$1.21 &
  56.78$\pm$0.61 &
  66.36$\pm$0.31 &
  25.14$\pm$0.24 &
  19.54$\pm$0.67 \\
\textit{+TA+WU} &
  \textbf{51.59$\pm$0.24} &
  \textbf{62.37$\pm$0.59} &
  16.66$\pm$0.72 &
  15.56$\pm$1.62 &
  \textbf{57.82$\pm$0.37} &
  \textbf{66.60$\pm$0.14} &
  19.86$\pm$0.60 &
  16.32$\pm$0.62 \\ \midrule
TKD &
  \textbf{53.73$\pm$0.25} &
  62.91$\pm$0.44 &
  20.77$\pm$0.58 &
  20.30$\pm$0.80 &
  58.90$\pm$1.02 &
  67.49$\pm$0.42 &
  17.38$\pm$0.80 &
  14.15$\pm$0.13 \\
\textit{+TA} &
  53.29$\pm$0.04 &
  \textbf{63.04$\pm$0.30} &
  \textbf{18.28$\pm$0.72} &
  \textbf{17.26$\pm$0.89} &
  58.80$\pm$0.48 &
  67.00$\pm$0.23 &
  \textbf{15.07$\pm$0.66} &
  \textbf{13.48$\pm$0.23} \\
\textit{+WU} &
  52.14$\pm$0.16 &
  60.83$\pm$0.24 &
  22.90$\pm$0.70 &
  25.45$\pm$0.55 &
  \textbf{59.85$\pm$0.95} &
  \textbf{67.77$\pm$0.46} &
  18.40$\pm$0.63 &
  16.35$\pm$0.64 \\
\textit{+TA+WU} &
  51.69$\pm$0.19 &
  61.09$\pm$0.28 &
  21.68$\pm$0.15 &
  23.15$\pm$0.59 &
  58.45$\pm$0.41 &
  66.94$\pm$0.26 &
  18.31$\pm$0.51 &
  16.40$\pm$0.60 \\ \midrule
ANCL &
  54.50$\pm$0.28 &
  63.96$\pm$0.32 &
  20.26$\pm$0.36 &
  17.56$\pm$0.40 &
  58.24$\pm$1.28 &
  66.98$\pm$0.70 &
  16.34$\pm$0.36 &
  11.40$\pm$0.48 \\
\textit{+TA} &
  54.39$\pm$0.64 &
  63.92$\pm$0.42 &
  18.04$\pm$0.33 &
  15.29$\pm$0.37 &
  59.16$\pm$0.63 &
  66.92$\pm$0.56 &
  15.03$\pm$0.91 &
  11.91$\pm$0.10 \\
\textit{+WU} &
  \textbf{54.84$\pm$0.20} &
  \textbf{64.25$\pm$0.37} &
  18.71$\pm$0.18 &
  16.55$\pm$0.51 &
  \textbf{60.14$\pm$0.66} &
  \textbf{67.78$\pm$0.41} &
  \textbf{14.04$\pm$0.27} &
  \textbf{10.56$\pm$0.42} \\
\textit{+TA+WU} &
  54.39$\pm$0.60 &
  64.08$\pm$0.40 &
  \textbf{17.81$\pm$0.53} &
  \textbf{15.00$\pm$0.59} &
  59.72$\pm$0.56 &
  67.00$\pm$0.20 &
  14.62$\pm$0.32 &
  12.12$\pm$0.51 \\ \midrule
 &
  \multicolumn{4}{c}{T10S10} &
  \multicolumn{4}{c}{T11S50} \\ 
  \midrule
GKD &
  40.33$\pm$0.37 &
  54.62$\pm$0.52 &
  32.72$\pm$0.09 &
  25.95$\pm$0.11 &
  43.34$\pm$0.85 &
  57.94$\pm$0.90 &
  26.04$\pm$2.37 &
  \textbf{14.47$\pm$0.83} \\
\textit{+TA} &
  \textbf{43.17$\pm$1.06} &
  \textbf{55.82$\pm$0.61} &
  25.10$\pm$1.03 &
  \textbf{20.52$\pm$0.24} &
  48.80$\pm$0.21 &
  57.18$\pm$0.45 &
  \textbf{17.56$\pm$0.53} &
  17.24$\pm$0.39 \\
\textit{+WU} &
  41.64$\pm$0.32 &
  53.12$\pm$0.61 &
  28.09$\pm$0.62 &
  27.10$\pm$0.55 &
  47.81$\pm$1.13 &
  \textbf{60.93$\pm$0.31} &
  23.24$\pm$1.20 &
  14.88$\pm$0.68 \\
\textit{+TA+WU} &
  42.37$\pm$1.01 &
  54.23$\pm$0.99 &
  \textbf{24.94$\pm$0.76} &
  23.31$\pm$0.65 &
  \textbf{50.79$\pm$0.11} &
  59.56$\pm$0.21 &
  19.53$\pm$0.13 &
  18.56$\pm$0.28 \\ \midrule
MKD &
  40.35$\pm$0.43 &
  54.01$\pm$0.01 &
  30.94$\pm$1.01 &
  28.19$\pm$0.61 &
  43.01$\pm$0.94 &
  56.18$\pm$0.90 &
  26.99$\pm$0.88 &
  14.94$\pm$0.17 \\
\textit{+TA} &
  42.35$\pm$0.50 &
  \textbf{56.02$\pm$0.20} &
  \textbf{19.37$\pm$1.18} &
  \textbf{18.60$\pm$0.76} &
  43.80$\pm$0.49 &
  52.05$\pm$0.24 &
  \textbf{13.67$\pm$1.09} &
  \textbf{14.23$\pm$0.46} \\
\textit{+WU} &
  40.49$\pm$0.52 &
  53.79$\pm$0.54 &
  29.89$\pm$1.41 &
  27.98$\pm$0.27 &
  43.61$\pm$1.63 &
  \textbf{57.32$\pm$1.60} &
  27.71$\pm$1.83 &
  16.09$\pm$3.04 \\
\textit{+TA+WU} &
  \textbf{42.94$\pm$0.88} &
  55.39$\pm$0.25 &
  20.62$\pm$0.53 &
  21.11$\pm$0.38 &
  \textbf{45.33$\pm$3.15} &
  54.05$\pm$3.32 &
  15.21$\pm$2.99 &
  15.60$\pm$2.54 \\ \midrule
TKD &
  43.19$\pm$0.16 &
  55.70$\pm$0.49 &
  24.84$\pm$0.35 &
  23.55$\pm$0.35 &
  40.56$\pm$1.30 &
  54.72$\pm$0.86 &
  15.39$\pm$1.01 &
  \textbf{10.16$\pm$0.34} \\
\textit{+TA} &
  \textbf{43.93$\pm$0.72} &
  \textbf{56.23$\pm$0.70} &
  \textbf{17.89$\pm$0.14} &
  \textbf{18.09$\pm$0.26} &
  42.83$\pm$0.61 &
  53.85$\pm$0.39 &
  \textbf{10.13$\pm$0.20} &
  13.15$\pm$0.16 \\
\textit{+WU} &
  42.71$\pm$1.29 &
  53.51$\pm$0.56 &
  24.84$\pm$0.96 &
  26.84$\pm$0.28 &
  45.44$\pm$0.98 &
  \textbf{58.94$\pm$0.45} &
  23.85$\pm$1.16 &
  18.09$\pm$0.76 \\
\textit{+TA+WU} &
  43.05$\pm$0.55 &
  54.40$\pm$0.64 &
  21.32$\pm$0.73 &
  22.50$\pm$0.12 &
  \textbf{49.48$\pm$0.92} &
  58.86$\pm$0.38 &
  17.65$\pm$1.12 &
  19.07$\pm$0.75 \\ \midrule
ANCL &
  44.01$\pm$0.19 &
  55.81$\pm$0.41 &
  29.56$\pm$0.51 &
  27.13$\pm$0.50 &
  46.94$\pm$0.67 &
  60.97$\pm$0.62 &
  26.30$\pm$0.84 &
  15.72$\pm$0.14 \\
\textit{+TA} &
  45.75$\pm$0.90 &
  \textbf{57.41$\pm$0.22} &
  23.71$\pm$1.08 &
  \textbf{21.54$\pm$0.50} &
  50.35$\pm$0.45 &
  59.22$\pm$0.26 &
  18.07$\pm$1.14 &
  17.34$\pm$0.53 \\
\textit{+WU} &
  45.08$\pm$0.08 &
  56.13$\pm$0.40 &
  27.27$\pm$0.34 &
  26.13$\pm$0.20 &
  50.18$\pm$0.87 &
  \textbf{63.15$\pm$0.12} &
  23.10$\pm$0.73 &
  \textbf{13.29$\pm$0.35} \\
\textit{+TA+WU} &
  \textbf{46.37$\pm$0.36} &
  57.20$\pm$0.32 &
  \textbf{22.79$\pm$0.74} &
  21.70$\pm$0.29 &
  \textbf{52.70$\pm$0.25} &
  61.18$\pm$0.34 &
  \textbf{17.93$\pm$0.82} &
  16.92$\pm$0.26 \\ \midrule 
 &
  \multicolumn{4}{c}{T20S5} &
  \multicolumn{4}{c}{T26S50} \\ 
  \midrule
GKD &
  24.14$\pm$0.91 &
  42.82$\pm$0.58 &
  45.53$\pm$0.88 &
  35.39$\pm$0.88 &
  14.82$\pm$0.85 &
  21.91$\pm$0.06 &
  17.99$\pm$0.53 &
  9.29$\pm$0.69 \\
\textit{+TA} &
  31.89$\pm$1.63 &
  \textbf{45.88$\pm$0.79} &
  \textbf{27.74$\pm$1.03} &
  \textbf{23.25$\pm$0.62} &
  17.16$\pm$0.84 &
  22.31$\pm$0.64 &
  \textbf{12.71$\pm$1.11} &
  \textbf{11.28$\pm$0.98} \\
\textit{+WU} &
  28.27$\pm$1.19 &
  43.92$\pm$0.38 &
  37.93$\pm$1.38 &
  32.37$\pm$0.66 &
  17.85$\pm$0.27 &
  30.15$\pm$0.56 &
  23.05$\pm$1.17 &
  15.86$\pm$0.91 \\
\textit{+TA+WU} &
  \textbf{32.63$\pm$0.69} &
  45.75$\pm$0.61 &
  29.33$\pm$0.37 &
  25.37$\pm$0.04 &
  \textbf{24.90$\pm$0.24} &
  \textbf{34.19$\pm$0.28} &
  20.82$\pm$1.10 &
  18.97$\pm$1.23 \\ \midrule
MKD &
  26.27$\pm$1.28 &
  43.39$\pm$0.66 &
  40.22$\pm$1.04 &
  34.25$\pm$0.81 &
  15.83$\pm$1.18 &
  26.07$\pm$0.29 &
  29.85$\pm$0.95 &
  16.00$\pm$0.06 \\
\textit{+TA} &
  31.90$\pm$0.34 &
  \textbf{46.18$\pm$0.54} &
  \textbf{18.92$\pm$0.14} &
  \textbf{19.14$\pm$0.79} &
  16.12$\pm$0.25 &
  22.25$\pm$0.13 &
  \textbf{11.65$\pm$0.66} &
  \textbf{11.94$\pm$1.23} \\
\textit{+WU} &
  27.50$\pm$1.10 &
  43.73$\pm$0.43 &
  38.01$\pm$1.61 &
  33.47$\pm$0.76 &
  16.15$\pm$0.27 &
  \textbf{28.42$\pm$1.65} &
  32.35$\pm$3.78 &
  18.39$\pm$4.16 \\
\textit{+TA+WU} &
  \textbf{33.55$\pm$1.15} &
  46.12$\pm$0.64 &
  22.06$\pm$1.28 &
  21.56$\pm$0.80 &
  \textbf{19.85$\pm$6.72} &
  25.79$\pm$6.22 &
  13.35$\pm$3.06 &
  13.72$\pm$3.01 \\ \midrule
TKD &
  29.39$\pm$0.47 &
  44.75$\pm$0.28 &
  35.96$\pm$0.95 &
  32.16$\pm$0.14 &
  10.99$\pm$0.22 &
  19.32$\pm$0.23 &
  13.55$\pm$0.88 &
  \textbf{9.67$\pm$0.61} \\
\textit{+TA} &
  32.71$\pm$0.27 &
  \textbf{46.45$\pm$0.42} &
  \textbf{19.12$\pm$1.02} &
  \textbf{19.55$\pm$0.30} &
  13.90$\pm$0.52 &
  22.55$\pm$0.83 &
  \textbf{7.24$\pm$0.71} &
  9.96$\pm$0.28 \\
\textit{+WU} &
  30.79$\pm$0.60 &
  45.29$\pm$0.38 &
  33.50$\pm$0.33 &
  31.09$\pm$0.94 &
  15.48$\pm$1.06 &
  27.81$\pm$0.55 &
  29.99$\pm$0.49 &
  23.07$\pm$0.97 \\
\textit{+TA+WU} &
  \textbf{33.73$\pm$0.68} &
  \textbf{46.45$\pm$0.27} &
  23.74$\pm$0.63 &
  22.98$\pm$0.53 &
  \textbf{22.59$\pm$0.66} &
  \textbf{32.46$\pm$0.52} &
  21.63$\pm$1.15 &
  20.65$\pm$0.99 \\ \midrule
ANCL &
  28.83$\pm$0.71 &
  44.94$\pm$0.63 &
  41.59$\pm$1.13 &
  34.24$\pm$1.14 &
  17.92$\pm$0.24 &
  29.19$\pm$0.76 &
  28.50$\pm$0.64 &
  16.12$\pm$0.27 \\
\textit{+TA} &
  33.92$\pm$0.68 &
  47.49$\pm$0.35 &
  \textbf{25.89$\pm$0.82} &
  \textbf{22.61$\pm$0.40} &
  23.52$\pm$0.73 &
  29.64$\pm$0.50 &
  \textbf{15.06$\pm$0.78} &
  \textbf{14.67$\pm$0.84} \\
\textit{+WU} &
  31.66$\pm$0.44 &
  46.28$\pm$0.56 &
  37.07$\pm$0.77 &
  32.07$\pm$1.20 &
  19.79$\pm$0.47 &
  32.25$\pm$0.15 &
  28.02$\pm$0.46 &
  15.11$\pm$0.46 \\
\textit{+TA+WU} &
  \textbf{35.79$\pm$0.39} &
  \textbf{47.81$\pm$0.30} &
  27.14$\pm$0.10 &
  24.93$\pm$0.89 &
  \textbf{27.49$\pm$1.10} &
  \textbf{33.31$\pm$0.58} &
  15.23$\pm$0.58 &
  14.68$\pm$0.35 \\
  \bottomrule
\end{tabular}%
}
\caption{Additional results for ImageNet100}
\label{tab:app:imagenet100}
\end{table*}

\subsection{Fine-grained classification using full datasets}
\label{sec:app:fg_full}
We conduct experiments using the same datasets as in the \Cref{sec:exp:finegrained}, but without sampling classes, so each of the 6 datasets is treated as a single task. Additionally, we add the results for experiments with a reversed order of tasks to ensure that the ordering does not affect our results. We report final and incremental accuracy and forgetting in \Cref{tab:app:full_fg}. We observe that Teacher Adaptation (TA) improves the final accuracy, but applying warmup (WU) in this setting results in significant drops in performance regardless of the order of tasks.

\begin{table*}
\centering
\resizebox{0.9\textwidth}{!}{%
\begin{tabular}{@{}lrrrrrrrr@{}}
\toprule
 &
  \multicolumn{4}{c}{Basic order} &
  \multicolumn{4}{c}{Reverse order} \\ \cmidrule{2-9}
 &
  \multicolumn{1}{c}{$Acc_{Final} \uparrow$} &
  \multicolumn{1}{c}{$Acc_{Inc} \uparrow$} &
  \multicolumn{1}{c}{$Forg_{Final} \downarrow$} &
  \multicolumn{1}{c}{$Forg_{Inc} \downarrow$} &
  \multicolumn{1}{c}{$Acc_{Final} \uparrow$} &
  \multicolumn{1}{c}{$Acc_{Inc} \uparrow$} &
  \multicolumn{1}{c}{$Forg_{Final} \downarrow$} &
  \multicolumn{1}{c}{$Forg_{Inc} \downarrow$} \\ \midrule
GKD &
  26.71$\pm$0.53 &
  \textbf{46.74$\pm$0.94} &
  46.57$\pm$0.89 &
  \textbf{37.47$\pm$1.07} &
  26.61$\pm$1.07 &
  54.88$\pm$0.62 &
  52.71$\pm$1.01 &
  39.34$\pm$0.41 \\
\textit{+TA} &
  \textbf{31.16$\pm$0.75} &
  46.38$\pm$0.74 &
  44.09$\pm$0.68 &
  37.84$\pm$0.98 &
  \textbf{33.61$\pm$0.40} &
  \textbf{54.97$\pm$0.35} &
  41.89$\pm$0.64 &
  \textbf{36.63$\pm$0.58} \\
\textit{+WU} &
  23.96$\pm$1.53 &
  36.37$\pm$0.95 &
  \textbf{34.78$\pm$1.80} &
  46.72$\pm$1.22 &
  23.60$\pm$1.33 &
  43.02$\pm$0.56 &
  39.25$\pm$1.36 &
  51.69$\pm$0.76 \\
\textit{+TA+WU} &
  24.74$\pm$0.95 &
  37.74$\pm$1.03 &
  40.44$\pm$0.87 &
  44.85$\pm$0.74 &
  27.06$\pm$0.44 &
  44.68$\pm$0.39 &
  \textbf{37.26$\pm$0.77} &
  47.10$\pm$0.61 \\ \bottomrule
\end{tabular}%
}
\caption{Additional results for training on full fine-grained datasets in standard and reversed order.}
\label{tab:app:full_fg}
\end{table*}

\subsection{Fine-grained classification with reverse order of datasets}
We show results for the experiments conducted in \Cref{sec:exp:finegrained}, extended with additional metrics. Additionally, we add the results for the results with reversed order of tasks, as in \Cref{sec:app:fg_full}. Similar to the results in the main paper, the impact of our method is more visible in settings with a larger number of tasks.

\begin{table*}
\centering
\resizebox{0.9\textwidth}{!}{%
\begin{tabular}{@{}lrrrrrrrr@{}}
\toprule
 &
  \multicolumn{4}{c}{6 tasks, 20 classes each, base order} &
  \multicolumn{4}{c}{6 tasks, 20 classes each, reverse order} \\ \cmidrule{2-9}
 &
\multicolumn{1}{c}{$Acc_{Final} \uparrow$} &
  \multicolumn{1}{c}{$Acc_{Inc} \uparrow$} &
  \multicolumn{1}{c}{$Forg_{Final} \downarrow$} &
  \multicolumn{1}{c}{$Forg_{Inc} \downarrow$} &
  \multicolumn{1}{c}{$Acc_{Final} \uparrow$} &
  \multicolumn{1}{c}{$Acc_{Inc} \uparrow$} &
  \multicolumn{1}{c}{$Forg_{Final} \downarrow$} &
  \multicolumn{1}{c}{$Forg_{Inc} \downarrow$} \\ \midrule
GKD &
  38.12$\pm$2.00 &
  56.03$\pm$3.75 &
  45.18$\pm$3.15 &
  \textbf{30.23$\pm$0.72} &
  37.10$\pm$3.40 &
  \textbf{67.30$\pm$1.99} &
  55.82$\pm$4.80 &
  34.60$\pm$3.77 \\
\textit{+TA} &
  \textbf{43.11$\pm$0.64} &
  \textbf{57.24$\pm$1.79} &
  39.58$\pm$0.80 &
  33.12$\pm$1.76 &
  \textbf{46.97$\pm$3.94} &
  67.19$\pm$4.81 &
  41.20$\pm$4.55 &
  \textbf{33.94$\pm$7.78} \\
\textit{+WU} &
  31.56$\pm$6.58 &
  44.06$\pm$5.43 &
  42.48$\pm$1.08 &
  49.24$\pm$2.65 &
  36.70$\pm$3.30 &
  54.88$\pm$1.53 &
  42.92$\pm$4.58 &
  51.30$\pm$3.39 \\
\textit{+TA+WU} &
  42.85$\pm$2.34 &
  52.96$\pm$1.20 &
  \textbf{35.97$\pm$2.75} &
  38.77$\pm$2.45 &
  44.05$\pm$1.94 &
  58.92$\pm$2.47 &
  \textbf{38.27$\pm$2.75} &
  45.16$\pm$4.57 \\ \midrule
MKD &
  41.32$\pm$0.81 &
  \textbf{60.12$\pm$1.59} &
  40.30$\pm$3.98 &
  \textbf{26.42$\pm$1.60} &
  43.96$\pm$3.89 &
  \textbf{70.65$\pm$1.67} &
  42.83$\pm$5.11 &
  \textbf{25.52$\pm$3.43} \\
\textit{+TA} &
  39.93$\pm$1.95 &
  55.77$\pm$2.23 &
  36.22$\pm$1.07 &
  31.43$\pm$2.28 &
  \textbf{46.94$\pm$4.42} &
  67.15$\pm$4.48 &
  \textbf{37.01$\pm$6.67} &
  31.07$\pm$8.38 \\
\textit{+WU} &
  34.65$\pm$2.72 &
  48.01$\pm$1.97 &
  44.86$\pm$0.61 &
  47.07$\pm$2.40 &
  35.49$\pm$5.51 &
  56.83$\pm$2.79 &
  46.14$\pm$6.67 &
  48.15$\pm$5.95 \\
\textit{+TA+WU} &
  \textbf{44.91$\pm$2.51} &
  56.41$\pm$1.04 &
  \textbf{35.53$\pm$2.19} &
  34.47$\pm$0.71 &
  44.62$\pm$2.85 &
  61.25$\pm$3.28 &
  38.31$\pm$3.91 &
  41.48$\pm$6.24 \\ \midrule
TKD &
  38.70$\pm$3.61 &
  56.69$\pm$3.36 &
  44.99$\pm$1.04 &
  33.86$\pm$1.36 &
  40.23$\pm$4.35 &
  66.84$\pm$2.79 &
  50.63$\pm$5.56 &
  35.77$\pm$4.85 \\
\textit{+TA} &
  42.34$\pm$0.77 &
  \textbf{57.99$\pm$1.88} &
  39.97$\pm$0.52 &
  \textbf{33.79$\pm$1.80} &
  \textbf{46.86$\pm$5.55} &
  \textbf{67.18$\pm$5.28} &
  39.90$\pm$6.90 &
  \textbf{33.67$\pm$8.84} \\
\textit{+WU} &
  32.71$\pm$5.27 &
  45.64$\pm$5.00 &
  44.43$\pm$2.13 &
  49.06$\pm$3.58 &
  37.53$\pm$2.65 &
  56.12$\pm$1.34 &
  44.06$\pm$4.80 &
  50.70$\pm$3.59 \\
\textit{+TA+WU} &
  \textbf{43.04$\pm$1.30} &
  53.68$\pm$1.56 &
  \textbf{37.14$\pm$1.52} &
  39.73$\pm$2.92 &
  44.52$\pm$2.14 &
  59.37$\pm$2.39 &
  \textbf{38.32$\pm$2.53} &
  45.65$\pm$4.61 \\ \midrule
 &
  \multicolumn{4}{c}{12 tasks, 10 classes each, base order} &
  \multicolumn{4}{c}{12 tasks, 10 classes each, reverse order} \\ \cmidrule{2-9}
 &
\multicolumn{1}{c}{$Acc_{Final} \uparrow$} &
  \multicolumn{1}{c}{$Acc_{Inc} \uparrow$} &
  \multicolumn{1}{c}{$Forg_{Final} \downarrow$} &
  \multicolumn{1}{c}{$Forg_{Inc} \downarrow$} &
  \multicolumn{1}{c}{$Acc_{Final} \uparrow$} &
  \multicolumn{1}{c}{$Acc_{Inc} \uparrow$} &
  \multicolumn{1}{c}{$Forg_{Final} \downarrow$} &
  \multicolumn{1}{c}{$Forg_{Inc} \downarrow$} \\ \midrule
GKD &
  23.66$\pm$2.75 &
  44.03$\pm$1.53 &
  58.93$\pm$1.75 &
  42.70$\pm$2.32 &
  26.84$\pm$4.11 &
  51.68$\pm$2.03 &
  59.80$\pm$5.93 &
  47.19$\pm$2.43 \\
\textit{+TA} &
  \textbf{36.46$\pm$0.94} &
  \textbf{50.80$\pm$1.27} &
  43.53$\pm$1.96 &
  37.58$\pm$3.01 &
  41.15$\pm$5.50 &
  \textbf{59.51$\pm$4.40} &
  42.61$\pm$6.36 &
  \textbf{36.14$\pm$6.20} \\
\textit{+WU} &
  11.02$\pm$1.11 &
  29.93$\pm$2.68 &
  50.32$\pm$3.34 &
  54.06$\pm$0.60 &
  19.55$\pm$0.93 &
  39.99$\pm$0.23 &
  46.40$\pm$0.80 &
  54.24$\pm$0.27 \\
\textit{+TA+WU} &
  36.00$\pm$0.89 &
  49.46$\pm$2.09 &
  \textbf{35.49$\pm$1.90} &
  \textbf{35.81$\pm$2.02} &
  \textbf{44.79$\pm$4.28} &
  52.88$\pm$5.06 &
  \textbf{26.81$\pm$5.14} &
  40.33$\pm$3.03 \\ \midrule
MKD &
  25.35$\pm$4.36 &
  45.19$\pm$2.91 &
  55.04$\pm$2.71 &
  43.81$\pm$1.56 &
  31.81$\pm$5.29 &
  53.63$\pm$2.31 &
  52.38$\pm$6.99 &
  43.86$\pm$3.05 \\
\textit{+TA} &
  \textbf{35.10$\pm$1.92} &
  \textbf{51.16$\pm$1.84} &
  37.96$\pm$1.84 &
  \textbf{34.89$\pm$2.60} &
  41.95$\pm$5.36 &
  \textbf{60.11$\pm$3.87} &
  36.16$\pm$6.75 &
  \textbf{33.18$\pm$5.64} \\
\textit{+WU} &
  15.61$\pm$6.67 &
  32.93$\pm$3.65 &
  49.31$\pm$7.17 &
  54.64$\pm$1.52 &
  26.19$\pm$8.31 &
  42.21$\pm$1.31 &
  44.49$\pm$6.28 &
  54.20$\pm$1.67 \\
\textit{+TA+WU} &
  34.65$\pm$4.00 &
  50.40$\pm$3.57 &
  \textbf{36.79$\pm$1.04} &
  35.74$\pm$3.45 &
  \textbf{45.91$\pm$4.64} &
  54.05$\pm$4.55 &
  \textbf{27.74$\pm$4.92} &
  40.08$\pm$3.32 \\ \midrule
TKD &
  24.69$\pm$3.39 &
  46.28$\pm$1.42 &
  56.57$\pm$2.87 &
  43.38$\pm$0.82 &
  31.50$\pm$3.26 &
  54.13$\pm$2.24 &
  52.51$\pm$5.19 &
  43.47$\pm$2.21 \\
\textit{+TA} &
  35.06$\pm$1.42 &
  \textbf{51.66$\pm$1.63} &
  39.72$\pm$2.08 &
  35.43$\pm$2.75 &
  41.40$\pm$4.78 &
  \textbf{59.22$\pm$3.67} &
  37.10$\pm$5.97 &
  \textbf{34.27$\pm$5.88} \\
\textit{+WU} &
  15.04$\pm$1.20 &
  33.32$\pm$1.74 &
  50.10$\pm$1.89 &
  53.59$\pm$1.52 &
  26.10$\pm$4.04 &
  42.34$\pm$1.31 &
  43.50$\pm$3.29 &
  52.66$\pm$0.59 \\
\textit{+TA+WU} &
  \textbf{35.81$\pm$1.27} &
  50.12$\pm$2.31 &
  \textbf{34.73$\pm$0.52} &
  \textbf{34.64$\pm$1.99} &
  \textbf{43.91$\pm$5.45} &
  52.38$\pm$5.54 &
  \textbf{26.77$\pm$4.76} &
  40.35$\pm$3.60 \\
  \midrule
\textit{} &
  \multicolumn{4}{c}{24 tasks, 5 classes each, base order} &
  \multicolumn{4}{c}{24 tasks, 5 classes each, reverse order} \\ \cmidrule{2-9}
 &
\multicolumn{1}{c}{$Acc_{Final} \uparrow$} &
  \multicolumn{1}{c}{$Acc_{Inc} \uparrow$} &
  \multicolumn{1}{c}{$Forg_{Final} \downarrow$} &
  \multicolumn{1}{c}{$Forg_{Inc} \downarrow$} &
  \multicolumn{1}{c}{$Acc_{Final} \uparrow$} &
  \multicolumn{1}{c}{$Acc_{Inc} \uparrow$} &
  \multicolumn{1}{c}{$Forg_{Final} \downarrow$} &
  \multicolumn{1}{c}{$Forg_{Inc} \downarrow$} \\ \midrule
GKD &
  9.41$\pm$2.78 &
  30.14$\pm$4.00 &
  60.16$\pm$4.76 &
  51.95$\pm$2.28 &
  15.33$\pm$2.59 &
  33.65$\pm$0.76 &
  64.43$\pm$3.19 &
  57.18$\pm$2.13 \\
\textit{+TA} &
  26.92$\pm$2.31 &
  42.86$\pm$1.82 &
  44.80$\pm$3.25 &
  39.01$\pm$1.55 &
  29.52$\pm$2.33 &
  \textbf{48.34$\pm$1.51} &
  48.32$\pm$2.43 &
  41.58$\pm$3.30 \\
\textit{+WU} &
  7.24$\pm$1.03 &
  24.91$\pm$1.44 &
  43.78$\pm$2.98 &
  51.18$\pm$1.15 &
  6.15$\pm$1.91 &
  25.45$\pm$2.08 &
  39.88$\pm$3.29 &
  46.36$\pm$1.88 \\
\textit{+TA+WU} &
  \textbf{27.30$\pm$1.97} &
  \textbf{45.28$\pm$2.25} &
  \textbf{34.75$\pm$4.58} &
  \textbf{33.67$\pm$2.48} &
  \textbf{34.14$\pm$9.24} &
  42.65$\pm$1.21 & 
  \textbf{19.92$\pm$11.23} &
  \textbf{32.38$\pm$8.24} \\ \midrule
MKD &
  11.57$\pm$2.70 &
  32.61$\pm$0.55 &
  63.70$\pm$2.58 &
  55.91$\pm$1.77 &
  15.53$\pm$2.96 &
  34.85$\pm$1.40 &
  62.28$\pm$3.64 &
  56.57$\pm$2.11 \\
\textit{+TA} &
  \textbf{27.88$\pm$1.58} &
  45.49$\pm$1.82 &
  37.65$\pm$1.16 &
  34.48$\pm$0.89 &
  \textbf{31.82$\pm$1.51} &
  \textbf{49.47$\pm$1.16} &
  38.89$\pm$1.18 &
  35.98$\pm$2.07 \\
\textit{+WU} &
  7.52$\pm$4.24 &
  24.62$\pm$1.56 &
  44.85$\pm$6.75 &
  52.59$\pm$2.59 &
  4.57$\pm$3.36 &
  21.36$\pm$4.30 &
  40.82$\pm$4.75 &
  51.73$\pm$0.78 \\
\textit{+TA+WU} &
  26.71$\pm$2.29 &
  \textbf{45.80$\pm$2.48} &
  \textbf{33.52$\pm$4.66} &
  \textbf{32.48$\pm$2.15} &
  30.63$\pm$6.78 &
  41.09$\pm$4.04 &
  \textbf{20.99$\pm$4.34} &
  \textbf{31.64$\pm$3.87} \\ \midrule
TKD &
  9.99$\pm$0.47 &
  32.17$\pm$2.71 &
  60.29$\pm$3.39 &
  51.72$\pm$1.95 &
  16.28$\pm$2.59 &
  34.68$\pm$1.62 &
  59.14$\pm$4.12 &
  54.83$\pm$1.69 \\
\textit{+TA} &
  \textbf{26.24$\pm$3.62} &
  \textbf{43.95$\pm$2.60} &
  36.28$\pm$2.17 &
  33.28$\pm$1.86 &
  \textbf{30.81$\pm$1.61} &
  \textbf{48.95$\pm$0.95} &
  38.10$\pm$1.53 &
  35.45$\pm$1.59 \\
\textit{+WU} &
  3.79$\pm$1.21 &
  22.99$\pm$1.33 &
  36.45$\pm$3.12 &
  42.30$\pm$4.03 &
  8.62$\pm$4.87 &
  24.91$\pm$6.55 &
  39.22$\pm$7.44 &
  49.71$\pm$2.49 \\
\textit{+TA+WU} &
  20.24$\pm$4.08 &
  37.22$\pm$6.11 &
  \textbf{23.39$\pm$2.94} &
  \textbf{24.53$\pm$3.01} &
  25.21$\pm$3.57 &
  47.46$\pm$5.21 &
  \textbf{28.40$\pm$4.04} &
  \textbf{26.15$\pm$2.93} \\ 
  \bottomrule
\end{tabular}%
}
\caption{Additional results for the fine-grained classification datasets, extended by the experiments with reversed order of the datasets.}
\label{tab:app:fg_rev}
\end{table*}

\subsection{DomainNet}

\begin{table*}[t]
\centering
\resizebox{\textwidth}{!}{%
\begin{tabular}{@{}lrrrrrrrr@{}}
\toprule
 & \multicolumn{8}{c}{6 tasks 50 classes with $\lambda = 10$}
 \\ \cmidrule{2-9}
 &
  \multicolumn{4}{c}{Trained from scratch} &
  \multicolumn{4}{c}{Pre-trained on Imagenet}
  \\ \cmidrule{2-9}
 &
  \multicolumn{1}{c}{$Acc_{Final} \uparrow$} &
  \multicolumn{1}{c}{$Acc_{Inc} \uparrow$} &
  \multicolumn{1}{c}{$Forg_{Final} \downarrow$} &
  \multicolumn{1}{c}{$Forg_{Inc} \downarrow$} &
  \multicolumn{1}{c}{$Acc_{Final} \uparrow$} &
  \multicolumn{1}{c}{$Acc_{Inc} \uparrow$} &
  \multicolumn{1}{c}{$Forg_{Final} \downarrow$} &
  \multicolumn{1}{c}{$Forg_{Inc} \downarrow$} \\ \cmidrule{2-9}
GKD &
  3.43$\pm$0.45 &
  18.63$\pm$0.27 &
  \textbf{31.05$\pm$0.42} &
  \textbf{23.27$\pm$0.26} &
  27.69$\pm$0.09 &
  43.27$\pm$0.10 &
  \textbf{41.61$\pm$0.71} &
  \textbf{36.83$\pm$0.88} \\
\textit{+TA} &
  \textbf{6.73$\pm$0.82} &
  \textbf{19.55$\pm$0.42} &
  36.26$\pm$0.55 &
  27.22$\pm$0.21 &
  \textbf{30.12$\pm$0.23} &
  \textbf{43.52$\pm$0.17} &
  43.38$\pm$0.39 &
  38.33$\pm$0.41 \\
\textit{+TA+WU} &
  5.83$\pm$0.70 &
  18.68$\pm$0.36 &
  33.00$\pm$0.59 &
  31.12$\pm$0.38 &
  28.83$\pm$0.26 &
  39.00$\pm$0.12 &
  42.64$\pm$0.56 &
  49.44$\pm$0.37 \\ 
\midrule
ANCL &
  7.33$\pm$0.91  & 19.58$\pm$0.46 & \textbf{33.28$\pm$0.18} & \textbf{25.63$\pm$0.20} & 25.58$\pm$0.62 & \textbf{42.90$\pm$0.84} & 45.01$\pm$1.19 & 37.28$\pm$1.86 \\
\textit{+TA} &
  8.85$\pm$0.25  & 20.34$\pm$0.40 & 39.70$\pm$0.38 & 30.73$\pm$0.44 & 26.95$\pm$0.07 & 42.67$\pm$0.51 & 41.62$\pm$0.52 & 38.56$\pm$1.24 \\
\textit{+TA+WU} &
  \textbf{9.65$\pm$0.40} & \textbf{20.39$\pm$0.09} & 37.57$\pm$0.12 & 32.83$\pm$0.66 & \textbf{29.15$\pm$0.98} & 42.44$\pm$0.80 & \textbf{34.23$\pm$0.85} & \textbf{37.09$\pm$1.33} \\

  \midrule
  & \multicolumn{8}{c}{12 tasks 25 classes with $\lambda = 10$}
  \\ \cmidrule{2-9}
  &
  \multicolumn{4}{c}{Trained from scratch} &
  \multicolumn{4}{c}{Pre-trained on Imagenet}
  \\ \cmidrule{2-9}
  &
  \multicolumn{1}{c}{$Acc_{Final} \uparrow$} &
  \multicolumn{1}{c}{$Acc_{Inc} \uparrow$} &
  \multicolumn{1}{c}{$Forg_{Final} \downarrow$} &
  \multicolumn{1}{c}{$Forg_{Inc} \downarrow$} &
  \multicolumn{1}{c}{$Acc_{Final} \uparrow$} &
  \multicolumn{1}{c}{$Acc_{Inc} \uparrow$} &
  \multicolumn{1}{c}{$Forg_{Final} \downarrow$} &
  \multicolumn{1}{c}{$Forg_{Inc} \downarrow$} \\ \cmidrule{2-9}
  GKD &
  2.48$\pm$0.41 &
  14.45$\pm$0.25 &
  29.01$\pm$0.63 &
  \textbf{29.04$\pm$0.37} &
  17.69$\pm$0.84 &
  35.98$\pm$0.96 &
  50.12$\pm$1.00 &
  43.00$\pm$1.66 \\
\textit{+TA} &
  4.88$\pm$0.46 &
  \textbf{16.25$\pm$0.46} &
  36.47$\pm$1.06 &
  33.03$\pm$0.19 &
  \textbf{28.02$\pm$0.68} &
  \textbf{38.89$\pm$0.52} &
  40.87$\pm$0.64 &
  \textbf{41.10$\pm$0.42} \\
\textit{+TA+WU} &
  \textbf{6.37$\pm$1.39} &
  16.22$\pm$0.45 &
  \textbf{27.45$\pm$0.70} &
  32.63$\pm$0.11 &
  27.97$\pm$0.38 &
  34.24$\pm$0.42 &
  \textbf{30.86$\pm$0.33} &
  44.17$\pm$0.37
 \\ \midrule
 ANCL &
  3.97$\pm$0.50  & 14.82$\pm$0.41 & \textbf{32.80$\pm$0.61} & \textbf{33.46$\pm$0.40} & 14.13$\pm$1.35 & 33.34$\pm$0.55 & 55.72$\pm$0.16 & 49.05$\pm$0.65 \\
\textit{+TA} &
  6.54$\pm$0.51  & 17.19$\pm$0.06 & 39.77$\pm$0.76 & 37.40$\pm$0.38 & 23.42$\pm$1.03 & 35.81$\pm$0.18 & 40.86$\pm$0.85 & 43.84$\pm$0.23 \\
\textit{+TA+WU} &
  \textbf{8.06$\pm$0.47} & \textbf{18.12$\pm$0.07} & 36.07$\pm$0.13 & 36.71$\pm$0.05 & \textbf{27.83$\pm$1.36} & \textbf{36.73$\pm$1.78} & \textbf{31.00$\pm$0.39} & \textbf{39.29$\pm$0.49} \\
  
 \midrule
 
 & \multicolumn{8}{c}{6 tasks 50 classes with $\lambda = 1$}
 \\ \cmidrule{2-9}
 &
  \multicolumn{4}{c}{Trained from scratch} &
  \multicolumn{4}{c}{Pre-trained on Imagenet}
  \\ \cmidrule{2-9}
 &
  \multicolumn{1}{c}{$Acc_{Final} \uparrow$} &
  \multicolumn{1}{c}{$Acc_{Inc} \uparrow$} &
  \multicolumn{1}{c}{$Forg_{Final} \downarrow$} &
  \multicolumn{1}{c}{$Forg_{Inc} \downarrow$} &
  \multicolumn{1}{c}{$Acc_{Final} \uparrow$} &
  \multicolumn{1}{c}{$Acc_{Inc} \uparrow$} &
  \multicolumn{1}{c}{$Forg_{Final} \downarrow$} &
  \multicolumn{1}{c}{$Forg_{Inc} \downarrow$} \\ \cmidrule{2-9}
GKD &
  5.52$\pm$2.11 &
  18.80$\pm$0.75 &
  \textbf{39.48$\pm$1.12} &
  \textbf{32.74$\pm$0.60} &
  19.93$\pm$0.47 &
  36.56$\pm$0.12 &
  58.83$\pm$0.96 &
  56.75$\pm$0.89
   \\
\textit{+TA} &
  \textbf{11.42$\pm$0.30} &
  \textbf{20.84$\pm$0.24} &
  42.92$\pm$0.49 &
  34.98$\pm$0.19 &
  \textbf{25.50$\pm$0.86} &
  \textbf{39.01$\pm$0.60} &
  \textbf{56.75$\pm$0.90} &
  \textbf{54.99$\pm$0.93} \\ \midrule
TKD &
  4.87$\pm$0.97 &
  18.94$\pm$0.36 &
  \textbf{32.95$\pm$0.78} &
  \textbf{26.60$\pm$0.29} &
  27.44$\pm$0.15 &
  42.18$\pm$0.41 &
  42.47$\pm$0.65 &
  \textbf{41.23$\pm$1.05} \\
\textit{+TA} &
  \textbf{6.76$\pm$1.03} &
  \textbf{19.67$\pm$0.19} &
  37.43$\pm$0.57 &
  29.91$\pm$0.09 &
  28.96$\pm$0.32 &
  \textbf{42.51$\pm$0.43} &
  43.91$\pm$0.16 &
  42.00$\pm$0.59 \\
\textit{+TA+WU} &
  5.88$\pm$0.82 &
  18.65$\pm$0.27 &
  33.50$\pm$0.96 &
  32.33$\pm$0.49 &
  \textbf{29.55$\pm$0.30} &
  40.62$\pm$0.07 &
  \textbf{42.08$\pm$0.55} &
  46.86$\pm$0.39
  \\ \midrule

MKD & 3.24$\pm$1.10  & 18.74$\pm$0.52 & \textbf{26.83$\pm$0.83} & \textbf{19.10$\pm$0.40}  & \textbf{30.41$\pm$0.47} & \textbf{45.70$\pm$0.30}   & 32.96$\pm$0.29 & \textbf{27.48$\pm$0.60}  \\
\textit{+TA} & 3.99$\pm$0.38 & 18.04$\pm$0.16 & 29.16$\pm$0.29 & 22.70$\pm$0.25 & 27.47$\pm$0.22 & 42.91$\pm$0.04 & \textbf{31.05$\pm$0.31} & 29.43$\pm$0.04 \\
\textit{+TA+WU} & \textbf{6.26$\pm$0.02} & \textbf{19.18$\pm$0.13} & 31.36$\pm$1.04 & 27.81$\pm$0.49 & 30.10$\pm$1.07 & 42.20$\pm$0.65 & 34.50$\pm$0.41 & 38.01$\pm$0.40

  \\ \midrule

 & \multicolumn{8}{c}{12 tasks 25 classes with $\lambda=1$}
 \\ \cmidrule{2-9}
 &
  \multicolumn{4}{c}{Trained from scratch} &
  \multicolumn{4}{c}{Pre-trained on Imagenet}
  \\ \cmidrule{2-9}
 &
  \multicolumn{1}{c}{$Acc_{Final} \uparrow$} &
  \multicolumn{1}{c}{$Acc_{Inc} \uparrow$} &
  \multicolumn{1}{c}{$Forg_{Final} \downarrow$} &
  \multicolumn{1}{c}{$Forg_{Inc} \downarrow$} &
  \multicolumn{1}{c}{$Acc_{Final} \uparrow$} &
  \multicolumn{1}{c}{$Acc_{Inc} \uparrow$} &
  \multicolumn{1}{c}{$Forg_{Final} \downarrow$} &
  \multicolumn{1}{c}{$Forg_{Inc} \downarrow$} \\ \cmidrule{2-9}
GKD &
  2.88$\pm$0.25 &
  14.38$\pm$0.82 &
  \textbf{38.89$\pm$3.42} &
  \textbf{39.78$\pm$2.23} &
  9.62$\pm$0.56 &
  27.99$\pm$0.40 &
  64.74$\pm$0.21 &
  59.93$\pm$1.13
   \\
\textit{+TA} &
  \textbf{6.80$\pm$0.28} &
  \textbf{17.02$\pm$0.19} &
  47.37$\pm$0.82 &
  43.44$\pm$0.16 &
  \textbf{19.00$\pm$0.83} &
  \textbf{33.90$\pm$0.62} &
  \textbf{58.20$\pm$0.71} &
  \textbf{54.97$\pm$0.71} \\ \midrule
TKD &
  3.39$\pm$0.43 &
  12.28$\pm$0.99 &
  41.22$\pm$0.28 &
  33.32$\pm$0.33 &
  22.66$\pm$1.59 &
  37.56$\pm$1.03 &
  41.65$\pm$1.73 &
  41.21$\pm$1.97 \\
\textit{+TA} &
  5.10$\pm$0.47 &
  \textbf{16.84$\pm$0.42} &
  34.03$\pm$0.61 &
  33.46$\pm$0.41 &
  \textbf{29.39$\pm$0.37} &
  \textbf{38.85$\pm$0.31} &
  34.51$\pm$0.54 &
  \textbf{39.91$\pm$0.39} \\
\textit{+TA+WU} &
  \textbf{6.27$\pm$0.52} &
  16.40$\pm$0.56 &
  \textbf{26.51$\pm$0.57} &
  \textbf{32.87$\pm$0.34} &
  29.26$\pm$2.05 &
  36.81$\pm$0.97 &
  \textbf{30.00$\pm$1.48} &
  41.98$\pm$0.95 \\ 
  
  \midrule
MKD & 2.32$\pm$0.39 & 13.45$\pm$0.53 & 23.92$\pm$1.02 & \textbf{27.23$\pm$0.98} & 24.55$\pm$0.20  & \textbf{39.14$\pm$0.21} & 38.09$\pm$1.02 & 36.53$\pm$1.00  \\
\textit{+TA} & 3.60$\pm$0.51  & 15.30$\pm$0.35  & 27.32$\pm$0.37 & 28.71$\pm$0.26 & 27.49$\pm$0.94 & 37.84$\pm$0.35 & 27.99$\pm$1.16 & \textbf{34.09$\pm$0.39} \\
\textit{+TA+WU} & \textbf{5.46$\pm$1.20} & \textbf{15.89$\pm$0.96} & \textbf{22.23$\pm$1.98} & 30.62$\pm$0.48 & \textbf{30.34$\pm$1.32}   & 37.79$\pm$1.11  & \textbf{24.81$\pm$1.05}  & 37.00$\pm$1.84 
  \\
  \bottomrule
\end{tabular}%
}
\caption{Additional results for DomainNet.}
\label{tab:app:domainnet}
\end{table*}

We present the full experimental results for DomainNet in \Cref{tab:domainnet_main}. In addition to incremental accuracy and forgetting, we also add final accuracy and forgetting to the metrics in the experiments. Additionally, we analyze the random initialization and impact of starting from pretrained checkpoint on ImageNet. For task-wise knowledge distillation (TKD) we apply $\lambda$ equal to 1, for global knowledge distillation we notice that $\lambda = 10$ works better. We set the learning rate to $0.01$ and use the scheduler with LR decay after 60, 120 and 180 epochs.

\end{appendix}